%% file: paper_jrnl_compsoc.tex
\UseRawInputEncoding 
\documentclass[10pt,compsoc]{IEEEtran}
\ifCLASSOPTIONcompsoc
  \usepackage[nocompress]{cite}
\else
  \usepackage{cite}
\fi
\ifCLASSINFOpdf
\else
\fi
\usepackage{graphicx}
\usepackage{amsmath}
\usepackage{amssymb}
\usepackage{booktabs}
\usepackage{algorithm}
\usepackage{multirow}
\usepackage{array}
\usepackage{color}
\usepackage{algpseudocode}
\usepackage{rotating}
\usepackage{tabularx}
\usepackage{url}
\usepackage{hyperref}

\hypersetup{
    urlcolor=blue,
    }

\begin{document}
%
\title{Cost Function Unrolling in Unsupervised Optical Flow}
%
%
%
%

\author{Gal~Lifshitz
        and~Dan~Raviv
\IEEEcompsocitemizethanks{
\IEEEcompsocthanksitem 
The authors are with Tel Aviv University, Tel Aviv-Yafo 69978, Israel (e-mail:
lifshitz@mail.tau.ac.il; darav@tauex.tau.ac.il).
}
}
\IEEEtitleabstractindextext{%
\input{sections/1_abstract}


}

\maketitle

\IEEEdisplaynontitleabstractindextext

%
\IEEEpeerreviewmaketitle


%
%
%
%

\input{sections/2_introduction}
\input{sections/3_related_work}

\input{sections/4_approach}

\input{sections/5_experimenation}
\input{sections/6_conclusions}

\ifCLASSOPTIONcompsoc
  \section*{Acknowledgments}
\else
  \section*{Acknowledgment}
\fi

This work is partially funded by the Yitzhak and Chaya Weinstein Research Institute for Signal Processing.

\ifCLASSOPTIONcaptionsoff
  \newpage
\fi



%

\bibliographystyle{IEEEtran}
\bibliography{egbib}


%

%
\input{bios/gal}
\input{bios/dan}






\end{document}

%% file: sections/1_abstract.tex
\begin{abstract}
Steepest descent algorithms, which are commonly used in deep learning, use the gradient as the descent direction, either as-is or after a direction shift using preconditioning. 
In many scenarios calculating the gradient is numerically hard due to complex or non-differentiable cost functions, specifically next to singular points. This has been commonly overcome by increased DNN model sizes and complexity.
In this work we propose a novel mechanism we refer to as Cost Unrolling, for improving the ability of a given DNN model to solve a complex cost function, without modifying its architecture or increasing computational complexity.
We focus on the derivation of the Total Variation (TV) smoothness constraint commonly used in unsupervised cost functions.
We introduce an iterative differentiable alternative to the TV smoothness constraint, which is demonstrated to produce more stable gradients during training, enable faster convergence and improve the predictions of a given DNN model.
We test our method in several tasks, including image denoising and unsupervised optical flow.
Replacing the TV smoothness constraint with our loss during DNN training, we report improved results in all tested scenarios. 
Specifically, our method improves flows predicted at occluded regions, a crucial task by itself, resulting in sharper motion boundaries.

\end{abstract}

%% file: sections/2_introduction.tex
\IEEEraisesectionheading{\section{Introduction}}
\IEEEPARstart{T}{he}
$L^1$ norm of the gradients of a given function, also known as its Total Variation (TV) semi-norm, and more specifically its estimation, has been a significant field of study in robust statistics \cite{Huber.Wiley.ea1981Robuststatistics,bredies2010total}. Even prior to the sweeping AI era, many approaches to Computer Vision problems, such as image restoration, denoising and registration \cite{RUDIN1992259,doi:10.1080/00207160500069904,zach2007duality}, have used a TV regularizer, as it represents the prior distribution of pixel intensities of natural images \cite{786990}. Its main  advantage is its robustness to small oscillations such as noise while preserving sharp discontinuities such as edges.

Historically, solving the TV problem has been a challenging task, mainly due to the non-differenetiability of the $L^1$ norm at zero. 
As a result, differentiable relaxations for $L^1$ have been proposed, such as the Huber \cite{huber1964robust} and Charbonnier \cite{charbonnier1997deterministic} loss functions. Further approaches involved iterative mechanisms which have demonstrated superior results in many tasks \cite{admm0,zach2007duality,RUDIN1992259}.
Introducing trainable Deep Neural Networks (DNNs) to Computer Vision has brought a significant performance boost, and specifically the commonly used auto-derivation frameworks \cite{pytorch,jia2014caffe,tensorflow2015-whitepaper} have provided quick and easy tools to solve complex functions. Indeed, these frameworks commonly bypass the $L^1$ non-differentiability either by differentiable relaxations or by simply using one of its sub-gradients, i.e.
\begin{equation} \label{eq:sub_gradient}
\begin{aligned}
    \frac{d|x|}{d x} =
    \begin{cases}
    1, & x\geq0  \\
    -1, & x<0
    \end{cases}
\end{aligned}
\end{equation}
which are highly discontinuous. Inspired by the vast research carried for many years before the deep learning era, we claim that non-differentiable cost functions should be dealt with greater care.
\input{figures/fig_teaser}

Proximal iterative methods for learning complex functions have shown superiority in many Image Processing, as well as Computer Vision tasks \cite{wang2016proximal,monga2019algorithm,zhang2020deep}. In these works, axiomatic optimization algorithms are unrolled and each iteration is mapped to a sub-network having its own learnable parameters. The resulting network architectures consist of task specific stacked modules which may be trained end-to-end in both supervised and unsupervised manners. Performance boost has been achieved either by increasing the number of stacked modules \cite{wang2016proximal} or introducing intermediate modules  \cite{luo2021upflow,wang2020cot}, both resulting in increased model complexity and size.

In this paper, to the best of our knowledge, we are the first to shift unrolling to the cost function, improving the optimization of a given DNN model while preserving its architecture, as opposed to baselines featuring unrolled architectures. 
Specifically, we present a novel pipeline we refer to as Cost Unrolling, in which 
we obtain a differentiable alternative for the the commonly used TV smoothness regularization.
Following the well known Alternating Direction Method of Multiplies (ADMM) \cite{admm0} algorithm, our initial optimization problem is decomposed into a set of iterative sub-problems, each one featuring a differentiable quadratic cost function. Our unrolled cost function is then defined as the accumulation of all unrolled sub-cost functions.
Gradients generated during training are shown to be much more stable in the regions where the gradients of the original cost function are hard to evaluate or undefined, improving convergence. 

We demonstrate the effectiveness of our unrolled cost function in synthetic experiments including image denoising, as well as the unsupervised optical flow task, which is the focus of this paper. The lack of available labeled data has made unsupervised optical flow learning a significant field of interest. 
Unsupervised training commonly consist of a measure for image consistency and TV smoothness regularization. 
Despite extensive research carried in the field, a major challenge which remains open is occlusions.
The occluded pixels, pixels within a reference image with no correspondence at the target image, are masked out when measuring image consistency. As a result, occluded regions, which are highly correlated to motion boundaries, are mostly dominated by the gradients of the smoothness constraint during model training. Interestingly, we find our method particularly beneficial at detecting motion in the occluded regions.

Our method is introduced to two recently published unsupervised optical flow baselines ARFlow \cite{liu2020learning} and SMURF \nolinebreak \cite{stone2021smurf}. Replacing their used TV smoothness constraint with our unrolled cost during all phases of training produces improved results on both MPI Sintel \cite{Butler:ECCV:2012} and KITTI 2015 \cite{Menze2015CVPR} unsupervised optical flow benchmarks.
Particularly, we report EPE reduced by up to 15.82\% on occluded pixels, where the smoothness constraint is dominant, allowing the detection of much sharper motion boundaries, as is highly visible in figures \ref{fig:teaser}, \ref{fig:occ_baselines} and \ref{fig:l1vsunroll}.

\subsection{Contributions}
We summarize our contributions as follows:
\begin{itemize}
    \item Present the Cost Unrolling mechanism for deep learning of TV regularized problems.
    \item Demonstrate our proposed pipeline in the image denoising and unsupervised optical flow domains.
    \item Report improved results on unsupervised optical flow benchmarks, achieved without modifying model architecture or increasing complexity.
\end{itemize}

%% file: figures/fig_teaser.tex
\begin{figure}
\begin{center}
\includegraphics[width=1\linewidth]{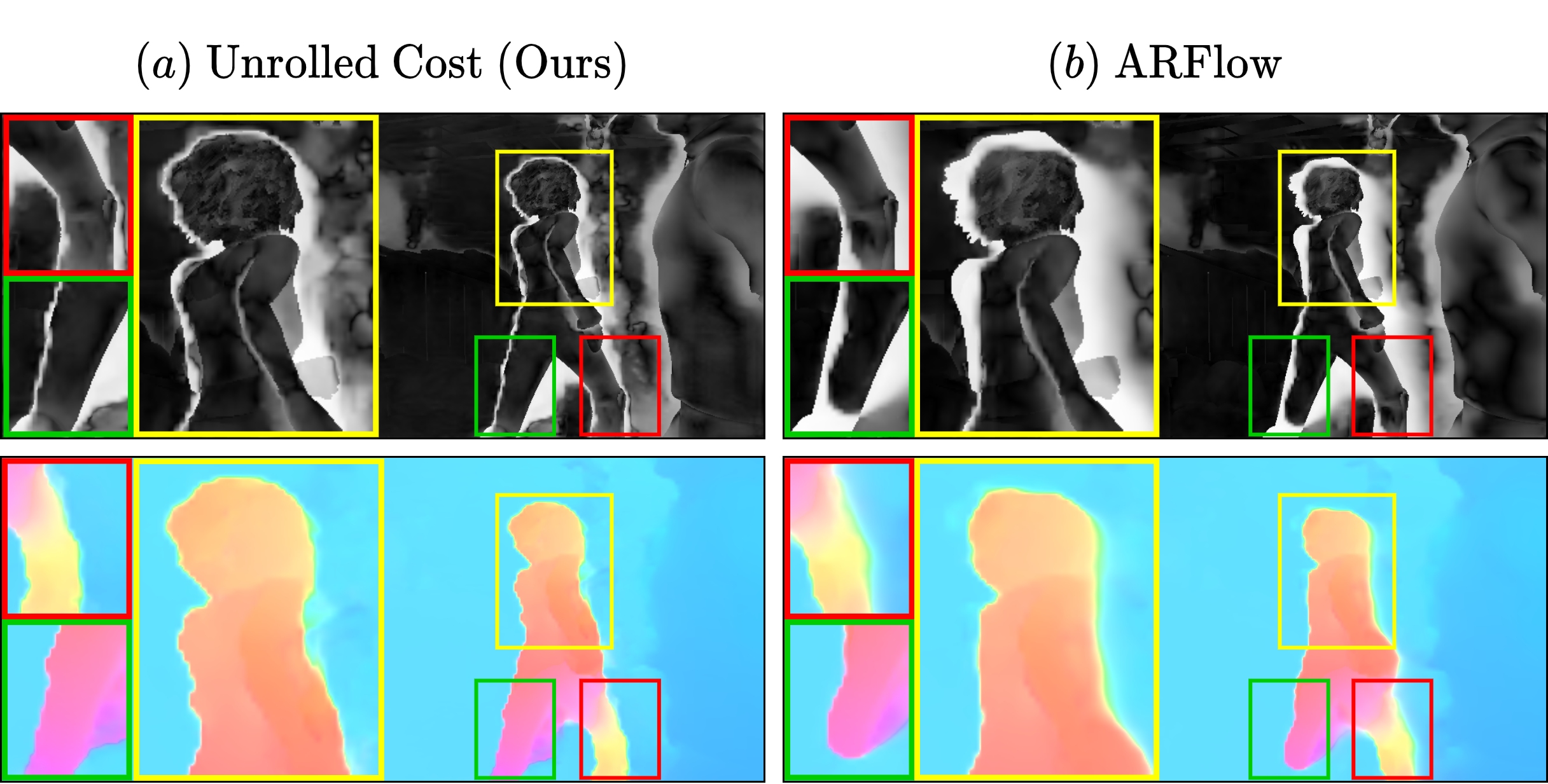}
\end{center}
   \caption{\textbf{Sintel Final benchmark qualitative example.}
   Training an optical flow model using our cost function enables the detection of sharper motion boundaries through improved convergence, without modifying the model's architecture or increasing complexity.
   Displayed are the predicted Sintel Final benchmark 'Market 1' flows (bottom) and errors (top) of both our method and the ARFlow \cite{liu2020learning} baseline, with close-ups on specific regions.
   White regions feature measured high errors.
   }
\label{fig:teaser}
\end{figure}

%% file: sections/3_related_work.tex
\section{Related Work}

Total Variation (TV) minimization has been a significant field of research in robust statistics \cite{Huber.Wiley.ea1981Robuststatistics}. It was introduced to Image Processing and Computer Vision tasks by \cite{RUDIN1992259} and showed superior performance over previous methods mainly due to its edge preserving property. However, solving TV regularized problems has shown to be non-trivial, mainly as the $L^1$ norm is non-differentiable at zero \cite{bredies2010total}. Differentiable relaxations were proposed by \cite{huber1964robust,charbonnier1997deterministic,Chambolle1997ImageRV}, however iterative methods have proven more effective \cite{doi:10.1080/00207160500069904,zach2007duality,admm0}.  A vast research regarding $L^1$ relaxations is given in \nolinebreak\cite{sun2014quantitative}.

Introducing trainable DNN models to Computer Vision has brought a significant performance boost to many tasks, such as image denoising \cite{gou2022multi}, graph matching \cite{lin2023graph} and optical flow \cite{teed2020raft}. The lack of labeled data available for training has pushed many DNN algorithms towards unsupervised learning. Not surprisingly, the TV smoothness regularization can be found in the cost functions of many unsupervised works both past and recent. As DNN algorithms use gradient descent algorithms for optimization, the non-differentiability of the $L^1$ norm at zero has been commonly bypassed either by differentiable approximations \cite{meister2017unflow} or by using its sub-gradients \nolinebreak (eq. \ref{eq:sub_gradient}) \cite{kim2020unsupervised,jonschkowski2020matters,luo2021upflow}.

Regularizer learning, a recent field of research, attempts to learn regularizers through incorporating designated learnable units. Learned analysis operators for regularized image restoration were proposed in \cite{chen2014insights}. Variational Networks, introduced in \cite{kobler2017variational}, featured a multi-residual \cite{abdi2016multi} structure inspired by the proximal gradient update steps.

Algorithm unrolling is an iterative structured approach for learning complex functions, in which iterative operations derived from axiomatic algorithms determine the model architecture. A vast review of the topic is presented in \cite{monga2019algorithm}. Axiomatic algorithms, such as the Alternating Direction Method of Multipliers (ADMM) \cite{admm0}, Half Quadratic Splitting (HQS) \cite{5445028} or Primal-Dual method \cite{chambolle2011first}, have inspired many deep algorithm unrolling frameworks. Indeed, ADMM-Net\nolinebreak \cite{sun2016deep} introduced a DNN model for compressed sensing with architecture derived from ADMM, USRNet\nolinebreak \cite{zhang2020deep} learned Single Image Super-Resolution (SISR) using a deep framework inspired by HQS and \cite{wang2016proximal} demonstrated an iterative structure derived from a Primal-Dual solver, aimed to refine the optical flow predicted by a FlowNet \cite{dosovitskiy2015flownet} model. 

Differentiable Programming (DP) is another approach for transferring axiomatic models to differentiable models, such as DNNs, demonstrating promising results. Indeed, \cite{peng2022xai} proposed unsupervised clustering using an NN model, and \cite{zhu2018dehazegan} proposed a learnable composition generator for single image dehazing, trained using an adversarial loss.

However, all of these frameworks feature closed form update steps derived from the learned objective function, which are mapped into task specific sub-networks, increasing model size and complexity. As opposed to previous works, we demonstrate improving the optimization of a given DNN model while preserving its architecture. We accumulate the iterative sub-objectives, unrolled following the ADMM algorithm \cite{admm0}, to obtain a differentiable alternative to the original non-differentiable TV smoothness regularization. While previous works designed dedicated unrolled learnable units, we present a pipeline that can be applied to any model architecture.

Learning in iterations has shown to be very effective at both image registration and optical flow tasks. While FlowNet and FlowNet 2.0 \cite{dosovitskiy2015flownet,ilg2017flownet} were the first to learn optical flow, they were surpassed in both supervised and unsupervised settings, by the well known PWC-Net \cite{Sun2018PWC-Net} which featured a course-to-fine iterative scheme. RAFT \cite{teed2020raft} introduced a fully iterative architecture which enabled a significant performance boost, as well as its expansion to the unsupervised settings SMURF \cite{stone2021smurf}. Further approaches, such as \cite{wang2016proximal,wang2020cot,luo2021upflow}, managed to improve the results of existing frameworks by adding intermediate modules and parameters for better flow refinement, which increase model size and complexity. 

\emph{Here we demonstrate that unlike all other methods, improved predictions of a given model can be achieved simply through improving the computed gradients during training.}

%% file: sections/4_approach.tex
\section{Cost Function Unrolling} \label{sec:approach}
A general formulation of the cost function used for unsupervised learning consists of a data term, measuring the likelihood of a given prediction over the given data, as well as a prior term, constraining probable predictions. In this work we consider the TV regularized unsupervised cost function and derive the iterative optimization steps for its minimization. These are then unrolled to obtain our novel smoothness regularizer. Denote by $\Theta$ the set of trainable parameters of a DNN model $ֿ\mathcal{F_\theta}$, its predicted output $\mathbf{F}$ and the set of unlabeled training data $\mathcal{I}$. The unsupervised TV regularized cost function takes the form:
\begin{equation} \label{eq:loss}
    \mathcal{L}(\mathbf{F},\mathcal{I}) = \mathbf{\Phi}\left( \mathbf{F},\mathcal{I} \right)+\lambda\|\nabla \mathbf{F} \|_1
\end{equation}
where $\mathbf{\Phi}$ is a differentiable function measuring the likelihood of $\mathbf{F}$ over the training data $\mathcal{I}$, $\nabla \mathbf{F}$ are its spatial gradients, $\|\cdot\|_1$ is the $L^1$ norm and $\lambda$ is a hyperparameter controlling regularization\footnote{For the rest of the discussion, we expand the $L^1$, $L^2$ norm and dot product to tensors $\mathbf{A},\mathbf{B}$ as $\|\mathbf{A}\|_1=\sum_i |A_i|$, $\|\mathbf{A}\|_2^2=\sum_i |A_i|^2$ and $\left<\mathbf{A},\mathbf{B} \right>_2=\sum_i A_i B_i$ where $A_i,B_i$ are corresponding elements of $\mathbf{A},\mathbf{B}$.}.

\subsection{Unrolling the Unsupervised Cost Function}
Our goal is to minimize the objective function in (\ref{eq:loss}). Following the ADMM \cite{admm0} algorithm, we derive the iterative update steps minimizing (\ref{eq:loss}).
Let us examine the following optimization problem:
\begin{equation} \label{eq: opt}
    \hat{\mathbf{F}}=\arg\min_\mathbf{F} \left\{ \mathbf{\Phi}\left( \mathbf{F},\mathcal{I} \right)+\lambda\|\nabla \mathbf{F} \|_1 \right\}
\end{equation}
We rewrite (\ref{eq: opt}), introducing an auxiliary variable $\mathbf{Q}$:
\begin{equation}
\begin{aligned}
    \mathbf{\hat{F},\hat{Q}}=&\arg\min_{\mathbf{F,Q}} \left\{ \mathbf{\Phi}\left( \mathbf{F},\mathcal{I} \right)+\lambda\|\mathbf{Q}\|_1 \right\} \\ &\text{s.t.} \quad \mathbf{Q=\nabla F}
\end{aligned}
\end{equation}
The augmented Lagrangian is then constructed as follows:
\begin{equation}
\begin{aligned}
     &\mathcal{L}_{\rho}(\mathbf{F,Q},\boldsymbol{\alpha}) = \mathbf{\Phi}\left( \mathbf{F},\mathcal{I} \right)+\lambda\|\mathbf{Q}\|_1\\ 
     &+ \mathbf{\left<\boldsymbol{\alpha},Q - \nabla F \right>}_2 + \frac{\rho}{2} \mathbf{\|Q - \nabla F\|}_2^2
\end{aligned}
\end{equation}
where $\boldsymbol{\alpha}$ is the Lagrange Multipliers tensor and $\rho$ is a penalty parameter. Substituting $\boldsymbol{\beta}=\frac{\boldsymbol{\alpha}}{\rho}$ for simplicity, we iteratively optimize $\mathbf{\{F,Q,\boldsymbol{\beta}\}}$ through solving the sub-problems (\ref{eq:admm_f}), (\ref{eq:admm_q}) and update the multipliers (\ref{eq:admm_b}) in each iteration:
\begin{subequations} 
\begin{align}
    \hat{\mathbf{F}}=&\min_{\mathbf{F}} \left\{\mathbf{\Phi}\left( \mathbf{F},\mathcal{I} \right) + \frac{\rho}{2} \mathbf{\|Q - \nabla F + \boldsymbol{\beta}\|}_2^2 \right\} \label{eq:admm_f} \\ 
    \hat{\mathbf{Q}}=&\min_{\mathbf{Q}} \left\{\mathbf{\lambda\|Q\|_1} + \frac{\rho}{2} \mathbf{\|Q - \nabla \hat{F} + \boldsymbol{\beta}\|}_2^2 \right\} \label{eq:admm_q} \\ 
    \hat{\boldsymbol{\beta}} = &\boldsymbol{\beta} + \mathbf{\left( Q - \nabla F \right)} \label{eq:admm_b}
\end{align}
\end{subequations}
The solutions to the problems in (\ref{eq:admm_f}) and (\ref{eq:admm_q}) are referred to as the ADMM update steps for $\mathbf{F}$ and $\mathbf{Q}$, respectively.

\input{algorithms/costunrolling}

\input{algorithms/training}

\input{figures/fig_unrolled_arch}

\subsection{Solving the Sub-Optimization Problems}
Solutions to the sub-optimization problems in equations (\ref{eq:admm_f}) and (\ref{eq:admm_q}) are now discussed. 
While the problem in equation (\ref{eq:admm_q}) has a closed form solution, deriving a closed form solution for the problem in (\ref{eq:admm_f}) is not trivial, as $\mathbf{\Phi}$ can be any function.
Substituting $\mathbf{C=\nabla F}$, (\ref{eq:admm_q}) reduces to:
\begin{equation}
\begin{aligned}
    &\min_{\mathbf{Q}} \left\{\mathbf{\lambda\|Q\|}_1 + \frac{\rho}{2} \mathbf{\|Q - C + \boldsymbol{\beta}\|}_2^2 \right\} \\
    &= \min_{\mathbf{Q}} \left\{\lambda \sum_{i}|Q_i| + \frac{\rho}{2} \sum_{i} |Q_i - C_i + \beta_i|^2 \right\} \\
    &= \sum_i \min_{Q_i} \left\{ \lambda |Q_i| + \frac{\rho}{2}|Q_i - C_i + \beta_i|^2 \right\}
    \label{eq:admm_qsol1}
    \end{aligned}
\end{equation}
i.e. the problem is separable and we may solve independently for each tensor element. The objective in (\ref{eq:admm_qsol1}) is convex and its solution is the well known Soft Thresholding operator $\hat{Q}_i = \mathcal{S}_{\lambda/\rho} \left( C_i-\beta_i \right)$, defined:
\begin{equation} \label{eq:S}
\begin{aligned}
    \mathcal{S}_{\lambda/\rho} (x) =
    \begin{cases}
    0, & |x| < \lambda / \rho \\
    x - \frac{\lambda}{\rho} \text{sign}(x), & |x| \geq \lambda / \rho
    \end{cases}
\end{aligned}
\end{equation}
We conclude the ADMM update step for $\mathbf{Q}$ (\ref{eq:admm_q}) as 
\begin{equation} \label{eq:admm_qsol2}
    \hat{\mathbf{Q}} = \mathcal{S}_{\lambda/\rho} \left( \mathbf{\nabla \hat{F}}-\boldsymbol{\beta}\right)
\end{equation}
where $\mathcal{S}_{\lambda/\rho}$ is applied element-wise. The Soft Thresholding operator performs shrinkage of the input signal, thus it promotes sparse solutions.

We now consider the sub-problem in equation (\ref{eq:admm_f}). 
Deriving a closed form solution is not trivial, as we wish to optimize for any $\mathbf{\Phi}$. However, note that the problem in \nolinebreak (\ref{eq:admm_f}) consists of the same data term as in (\ref{eq: opt}) with the TV smoothness regularizer replaced by a softer, differentiable constraint, denoted:
\begin{equation} \label{eq:ell}
    \ell(\mathbf{F}) = \frac{\rho}{2}\left\| \mathbf{Q} + \boldsymbol{\beta} - \nabla \mathbf{F} \right\|_2^2
\end{equation}
Following this realization, we construct an alternative for the objective function in (\ref{eq: opt}), which may optimized using our initial DNN model $\mathcal{F}_\theta$ thanks to its differentiability. We now elaborate on this process.



\subsection{Unrolled Cost Function}
Inspired by the ADMM update steps given in equations \nolinebreak (\ref{eq:admm_f}),(\ref{eq:admm_q}),(\ref{eq:admm_b}), our unrolled smoothness constraint consists of the normalized accumulation of all the iteratively generated constraints in the form of (\ref{eq:ell}), mainly:
\begin{equation}
\begin{aligned}
    \mathcal{L}_\text{sm}^T(\mathbf{F})&= \frac{1}{T} \sum_{t=1}^{T} \alpha_t \ell^{(t)}(\mathbf{F)} \\
    &=\frac{\rho}{2} \frac{1}{T} \sum_{t=1}^{T} \alpha_t \|\mathbf{Q}^{(t-1)} + \boldsymbol{\beta}^{(t-1)} - \nabla\mathbf{F} \|_2^2 \label{eq:loss_sm}
\end{aligned}
\end{equation}
where:
\begin{subequations} 
\begin{align}
    \mathbf{Q}^{(t)}&=\mathcal{S}_{\lambda/\rho} \left( \mathbf{\nabla F}-\boldsymbol{\beta}^{(t-1)} \right) \label{eq:update_q} \\ 
    \boldsymbol{\beta}^{(t)}&= \boldsymbol{\beta}^{(t-1)} + \left( \mathbf{Q}^{(t)} - \nabla \mathbf{F} \right) \label{eq:update_b}
\end{align}
\end{subequations}
$\mathbf{Q}^{(0)},\boldsymbol{\beta}^{(0)}$ are initialized to zeros, $T$ is a hyperparameter stating the number of update steps carried, and $\{\alpha_t\}$ adjust the weight of each step.

Suppose we limit the number of update steps to $T=2$. Following (\ref{eq:update_q}), (\ref{eq:update_b}) yields:
\begin{subequations} 
\begin{align}
    \mathbf{Q}^{(1)}&=\mathcal{S}_{\lambda/\rho} \left( \mathbf{\nabla F}-\boldsymbol{\beta}^{(0)} \right) =  \mathcal{S}_{\lambda/\rho} \left( \mathbf{\nabla F} \right)\label{eq:q1} \\ 
    \boldsymbol{\beta}^{(1)}&= \boldsymbol{\beta}^{(0)} + \left( \mathbf{Q}^{(1)} - \nabla \mathbf{F} \right) = \mathcal{S}_{\lambda/\rho} \left( \mathbf{\nabla F} \right) - \nabla \mathbf{F} \label{eq:b1}
\end{align}
\end{subequations}
which results in the following truncated smoothness constraint:
\begin{equation} 
    \mathcal{L}_\text{sm}^2(\mathbf{F})
    =\frac{\rho}{4}\left\{ \alpha_1\|\nabla\mathbf{F}\|_2^2 + 2\alpha_2 \|\mathcal{S}_{\lambda/\rho} \left(\nabla \mathbf{F} \right) - \nabla \mathbf{F}\|_2^2\right\} \label{eq:loss_sm_tr}
\end{equation}
Recall that minimizing the TV of a function promotes sparse output gradients. 
In fact, eq. (\ref{eq:loss_sm_tr}) can be viewed as Tikhonov regularization over $\nabla\mathbf{F}$, which implicitly yields a differentiable alternative for TV minimization. Specifically, note that its minimization involves the minimization of the $L^2$ distance between the true and sparsified output gradients. Thus, it promotes sparse output gradients. 
Experiments presented in sections \ref{sec:exp_steps} and \ref{sec:exp_comp} provide empirical support for the optimality of $T=2$, mainly as it manages to both promote sparse output gradients and maintain computational complexity.

Interestingly, our truncated constraint given in eq. (\ref{eq:loss_sm_tr}) can be considered as a further relaxation of the well known Huber loss \cite{huber1964robust}. The Huber loss function (given in eq. (\ref{eq:huber})) is essentially the Moreau envelope of the $L^1$ loss function:
\begin{equation}
\begin{aligned}
    r_H(\nabla\mathbf{F};k) &= \min_\mathbf{y}\| \mathbf{y} - \nabla\mathbf{F}\|_2^2 + k\|\mathbf{y}\|_1 \\
    &= \|\mathcal{S}_k \left(\nabla\mathbf{F} \right) - \nabla\mathbf{F}\|_2^2 + k\|\mathcal{S}_k \left(\nabla\mathbf{F} \right)\|_1
\end{aligned}
\end{equation}
The main difference between ours and the Huber loss functions is the $L^1$ norm replaced by $L^2$, gaining $L^2$ behavior everywhere as opposed to Huber.
We claim that this further relaxation featured in our method, inspired by the ADMM update steps, enables faster convergence and thus improves DNN training. 
We verify this empirically through rigorous experimentation in section \ref{sec:exp}.
\input{figures/fig_toy_graphs}

In conclusion, our modified differentiable cost function takes the form:
\begin{equation} \label{eq:loss_ours}
    \mathcal{L}_\text{dif}(\mathbf{F},\mathcal{I}) = \mathbf{\Phi}\left( \mathbf{F},\mathcal{I} \right) + \mathcal{L}_\text{sm}^T(\mathbf{F})
\end{equation}
Following ADMM, we have derived a quadratic regularizer, which consists of soft thresholding operations, thus essentially enforces $L^1$ minimization.
Note that here, instead of explicitly refining $\mathbf{F}$ in each iteration as suggested by the iterative ADMM update steps in (\ref{eq:admm_f}),(\ref{eq:admm_q}),(\ref{eq:admm_b}), we optimize all iteratively generated smoothness constraints simultaneously given a model prediction. Our proposed smoothness constraint computation and training scheme are summarized in algorithms \ref{alg:unrolling} and \ref{alg:training} respectively. A block diagram is given in figure \ref{fig:unrol_arch}. Note that our method affects only the training phase of a given DNN model, hence no complexity is added during inference. Interestingly, as demonstrated in section \nolinebreak \ref{sec:exp}, setting $T=2$, we manage to preserve time and memory consumption during training as well.

%% file: algorithms/costunrolling.tex
\begin{figure}
    \centering
    \begin{algorithm}[H]
    \caption{Unrolled Smoothness Constraint}\label{alg:unrolling}
    \begin{algorithmic}[1]
    \Require 
    \Statex Model prediction $\mathbf{F}$
    \Statex No. of cost function update steps $T$, $\{\alpha_t\}_{t=1}^T$
    \Statex Hyperparameters $\lambda$, $\rho$
    
    \Procedure{UnrolledSmooth}{$\mathbf{F}$,$T$,$\lambda$,$\rho$}
    \State $\boldsymbol{\beta}^{(0)}\gets\mathbf{0}$
    \State $\mathbf{Q}^{(0)}\gets\mathbf{0}$
    \State $\nabla \mathbf{F}\gets \textsc{ComputeGradient}(\mathbf{F})$

    \For {$t \leftarrow 1, T$} 
        \State $\ell^{(t)}(\mathbf{F}) \gets \frac{\rho}{2}\left\| \mathbf{Q}^{(t-1)} + \boldsymbol{\beta}^{(t-1)} - \nabla \mathbf{F} \right\|_2^2$
        \State $\mathbf{Q}^{(t)}\gets \mathcal{S}_{\lambda/\rho}\left( \nabla \mathbf{F} - \boldsymbol{\beta}^{(t-1)} \right)$ \Comment{Update $\mathbf{Q}$}
        \State $\boldsymbol{\beta}^{(t)} \gets \boldsymbol{\beta}^{(t-1)} + \mathbf{Q}^{(t)} - \nabla \mathbf{F}$  \Comment{Update $\boldsymbol{\beta}$}
    \EndFor
    \State $\mathcal{L}_\text{sm}^T(\mathbf{F}) \gets \frac{1}{T} \sum_{t=1}^T \alpha_t \ell^{(t)}(\mathbf{F})$
    \EndProcedure
    \Ensure Smoothness constraint $\mathcal{L}_\text{sm}^T (\mathbf{F})$
    \end{algorithmic}
    \end{algorithm}
\end{figure}

%% file: algorithms/training.tex
\begin{figure}
    \centering
    \begin{algorithm}[H]
    \caption{Model Training}\label{alg:training}
    \begin{algorithmic}[1]
    \Require 
    \Statex Model parameters $\theta$, initialized DNN model $\mathcal{F}_\theta$ 
    \Statex Set of training data $\mathcal{I}$ 
    \Statex Liklihood measure $\mathbf{\Phi}$
    \Statex Smoothness hyperparameters $T,\lambda,\rho$
    \Procedure{ModelTraining}{$\mathcal{F}_\theta$,$\mathcal{I}$,$\mathbf{\Phi}$}

    \For {$k \leftarrow 1, K$} \Comment{Training iterations}
        \State $\mathcal{B}\gets \{\mathbf{x}_i \sim \mathcal{I}\}_{i=1}^N$ \Comment{Sample data batch}
        \State $\mathbf{F} \gets \mathcal{F}_\theta\left(\mathcal{B}\right)$
        \Comment{Model Prediction}
        \Statex \emph{Loss Calculations:}
        \State $\mathcal{L}_\text{sm}^T\left( \mathbf{F} \right) \gets \textsc{UnrolledSmooth}(\mathbf{F},T,\lambda,\rho)$
        \State $\mathcal{L}_\text{dif}\left( \mathbf{F},\mathcal{B} \right) \gets \mathbf{\Phi}\left( \mathbf{F},\mathcal{B} \right) + \mathcal{L}_\text{sm}^T(\mathbf{F)}$ 
        \Statex \emph{Update Parameters:}
        \State $\theta \gets$optimizer$\left( \theta,\mathcal{L}_\text{dif} \right)$ 
    \EndFor
    \EndProcedure
    \Ensure Trained DNN model $\mathcal{F}_\theta$
    \end{algorithmic}
    \end{algorithm}
\end{figure}

%% file: figures/fig_unrolled_arch.tex
\begin{figure*}
\begin{center}
\includegraphics[trim=0 0 100 0,  width=1\linewidth]{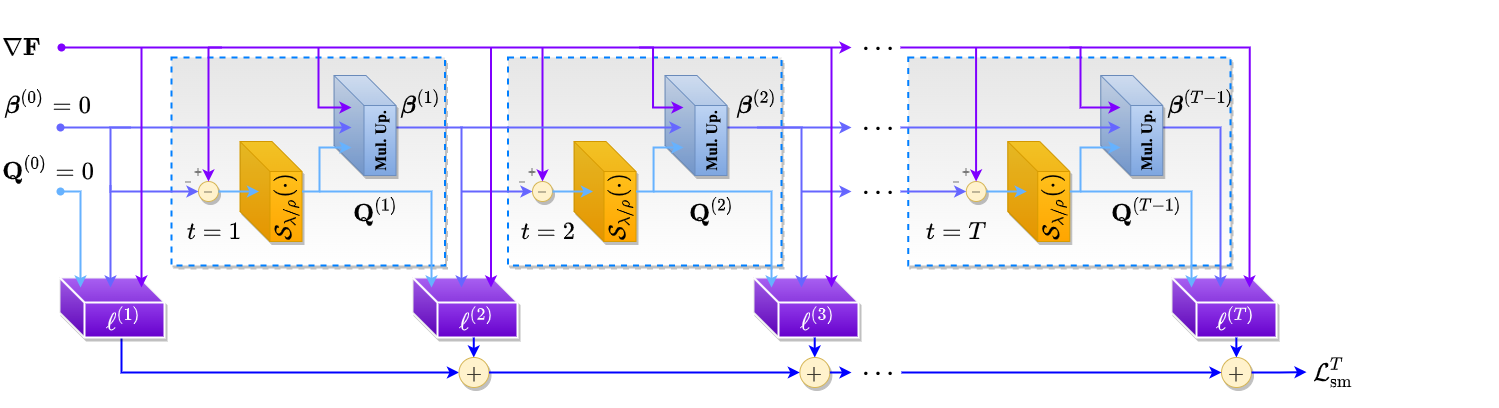}
   \caption{\textbf{Unrolled smoothness constraint block diagram.} 
   In each training iteration, given a flow prediction, its spatial gradient $\mathbf{\nabla F}$ is derived.
   Initialized at $\boldsymbol{\beta}^{(0)}=\mathbf{Q}^{(0)}=\mathbf{0}$, the Soft Thresholding and Multipliers Update steps are carried for update steps $t \in \{1,...,T\}$ to produce $\{\mathbf{Q}^{(t)},\boldsymbol{\beta}^{(t)} \}_{t=1}^{T}$, which are then used together with $\mathbf{\nabla F}$ to construct our smoothness constraint $\mathcal{L}_\text{sm}^T$ as in (\ref{eq:loss_sm}).}
   
   \label{fig:unrol_arch}
\end{center}
\end{figure*}

%% file: figures/fig_toy_graphs.tex
\begin{figure*}
\begin{center}
\includegraphics[trim=10 10 0 0, width=1\linewidth]{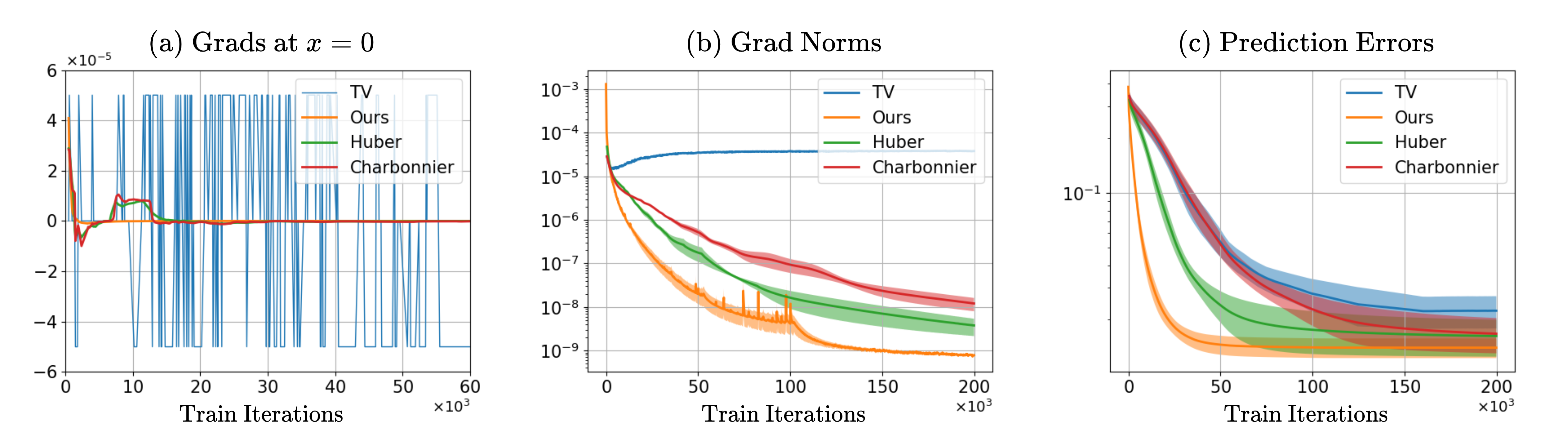}
    \caption{\textbf{PC signal prediction - training.} In each experiment we train a simple model to predict a full randomly generated PC 1D signal given a fraction of its samples and smoothness regularization. We train using our method, as well as standard TV and common $L^1$ relaxations.
    Displayed are mean and standard deviation of gradient norms (b) and prediction errors (c) measured during training while performing several independent experiments of the best configurations. An example of gradients recorded at a selected point is given in (a).}
   \label{fig:toy_graphs}
\end{center}
\end{figure*}

%% file: sections/5_experimenation.tex
\section{Experimentation} \label{sec:exp}

We study the effectiveness of our method through conducting experiments in three domains: piece-wise constant (PC) signal prediction (section \ref{sec:synthtetic}), image denoising (section \ref{sec:denoising}) and unsupervised optical flow (section \ref{sec:of}). All of our models are implemented in Pytorch \cite{pytorch}. \footnote{Our code is available at \url{https://github.com/gallif/CostFunctionUnrolling.git}.}

We further compare our unrolled cost to standard TV smoothness regularization, as well as $L^1$ relaxation baselines: Huber \cite{huber1964robust} and Charbonnier \cite{charbonnier1997deterministic} loss functions. 
The Huber regularizer over a prediction $\mathbf{x}$ is defined as $\mathcal{R}_H(\mathbf{x};k)=\sum_i r_H (x_i;k)$, where:
\begin{equation} \label{eq:huber}
\begin{aligned}
    r_H(x_i;k) =
    \begin{cases}
    \frac{1}{2}x_i^2, & |x_i| < k \\
    k|x_i| -\frac{1}{2}k^2 & |x_i| \geq k
    \end{cases}
\end{aligned}
\end{equation}
and $x_i$ are the elements of $\mathbf{x}$. It aims to take advantage of the $L^1$ norm robustness to outliers, but suggests better convergence replacing $L^1$ by $L^2$ for points near its optimum. The Charbonnier regularizer is defined as $\mathcal{R}_C(\mathbf{x})=\sum_i r_C(x_i)$, where:
\begin{equation} \label{eq:charb}
    r_C(x_i)=\sqrt{x_i^2+\epsilon^2}
\end{equation}
Following the discussion in \cite{sun2014quantitative}, we focus on small $k,\epsilon$ values. Note that both functions approach the $L^1$ norm only for $k,\epsilon\rightarrow0$, and resemble the $L^2$ norm otherwise. 
We show that our iterative approach is superior to these $L^1$ relaxations, achieving improved results and faster convergence.


\subsection{Piece-wise Constant (PC) Signal Prediction} \label{sec:synthtetic}
\input{tables/toy_results}
\input{figures/fig_img_denoising_all}

Firstly, we perform a series of experiments over synthetic 1D Piece-wise Constant (PC) signals in order to study the generated gradients and convergence of our smoothness constraint. In each experiment, we optimize a small NN model to predict a randomly generated full PC signal given only a fixed fraction of its samples, performing GD iterations until convergence. Each cost function consists of a data term, comparing the samples and corresponding predictions, as well as a smoothness regularizer. Note that although the data term is supervised, it contains knowledge of the original signal only at the sampled points. As a result, the full signal reconstruction is heavily dependent on the choice of smoothness term. In fact, these experiments are rather simplified 1D reductions of what is commonly known in unsupervised registration tasks as \emph{occluded regions} featuring no correspondence, making the smoothness constraint primary during training on these regions.

\subsubsection{Implementation}
We generate 1D PC signals $y(x),x\in[a,b]$ as linear combinations of random rectangular pulses, and sample them uniformly to generate our data $\{(x_i,y_i)\}_{i=0}^{N-1}$ where $y_i=y(x_i)$. In each experiment, we optimize a small NN model, consisting of 4 FC layers to predict the full signal given its samples, using GD iterations. Defining our data term $\mathbf{\Phi}\left( \mathbf{f},\mathbf{y} \right)=\frac{1}{N}\sum_{i=0}^{N-1} \left(f_i - y_i\right)^2$, where $\{f_i\}_{i=0}^{N-1}$ are the corresponding samples of the predicted output, our loss function takes the form:
\begin{equation}
\label{eq:toy_data}
    \mathcal{L}_{\text{dif}}\left( \mathbf{f},\mathbf{y} \right) = \frac{1}{N}\sum_{i=0}^{N-1} \left(f_i - y_i\right)^2 + \mathcal{L}^T_{\text{sm}}(\mathbf{f};\lambda,\rho)
\end{equation}
We optimize our model using our unrolled cost and compare to standard TV, Huber and Charbonnier, measuring the prediction error and the backpropagating gradients' norm during training. These enable studying and comparing the behavior of the tested loss functions. The prediction error is defined as $\frac{1}{L} \sum_{i=1}^L \left| f_i - y_i \right|$, where $f_i,y_i$ are the predicted and ground-truth full signals of length $L$, respectively.
The gradient's norm is computed as $\frac{1}{L}\sum_{i=1}^L \left|\frac{\partial \mathcal{L}}{\partial f_i}\right|$, where $\mathcal{L}$ is the overall loss function used for training.
In order to resemble actual training of an NN model, all gradients were computed using auto-differentiation.
We conduct several independent experiments, during which we record the backpropagating gradients as well as the prediction errors. Further visualizations as well as implementation details and hyperparameter examinations are given in Appendix \nolinebreak A (available online). 


\subsubsection{Results}
\input{tables/image_results_spec}

Table \ref{ta:toy} summarizes measured final gradient norms and prediction errors averaged over our conducted experiments of our best configurations. Corresponding gradient norms and prediction errors recorded during training, as well as en example of gradients recorded at a selected point are given in figures \ref{fig:toy_graphs}(b),(c),(a) respectively.
Recall that TV regularization promotes piece-wise constant solutions. In fact, one of its most notable shortcomings is the staircasing phenomenon \cite{bredies2010total}, in which solutions are random piece-wise constant signals featuring equal TV. This severe instability is caused by the discontinuous nature of the TV gradients generated during training.
Indeed, the staircasing phenomenon is highly evident in our TV experiments, in terms of severe gradient oscillations and non-decreasing gradient norm shown in figures \ref{fig:toy_graphs}(a),(b). More visualizations are given in Appendix A (available online).
On the other hand, our smoothness constraint manages to promote deterministic, smooth but edge preserving solutions, with its gradients rapidly decreasing to zero, enabling the fastest convergence.

An extensive hyperparameter study of our proposed method, as well as Huber and Charbonnier is given in Appendix A.2 (available online). The best configurations are summarized in figure \ref{fig:toy_graphs}, revealing perhaps the most significant benefit of our smoothness constraint.
Both Huber and Charbonnier essentially relax the $L^1$ loss function by modifying its shape around its optimum making it differentiable. Indeed, this enables gradients generated during training to decay towards reaching the optimum as opposed to TV, as is seen in figures \ref{fig:toy_graphs}(a),(b). Despite being made differentiable, training using the proposed $L^1$ relaxations still suffers from evident oscillations as a result of their constructed shapes, which result in slower convergence. 
Indeed, as shown in Appendix A.2 (available online), the convergence rate and errors of both Huber and Charbonnier vary heavily as function of the chosen $k,\epsilon$, as opposed to our method. We re-emphasize here that we are interested in small $k,\epsilon$ values.
Instead of letting a DNN model to minimize the difficult (perhaps relaxed) TV smoothness constraint, we de facto provide further guidance during training, through forcing $L^2$ similarity between the true and explicitly computed sparsified versions of its prediction gradients. 
In fact, explicitly computing the sparsified output gradients during the forward pass replaces the difficult gradient computation during backpropagation. This, together with the well behaved $L^2$ minimization, produces the most stable gradients during training, featuring a rapid decrement. 
As is demonstrated in figure \ref{fig:toy_graphs}, our smoothness constraint converges to its optimum more than two times faster than all other baselines.


\subsection{Image Denoising}
\label{sec:denoising}

Broadening the demonstration of our proposed method, we test our unrolled smoothness constraint in image denoising. We recreate several experiments from the well known Deep Image Prior (DIP) \cite{ulyanov2018deep} baseline, which were furtherly enhanced in \cite{liu2019image} by adding TV regularization (DIP-TV), particularly in low SNRs. The proposed baseline consists of training the weights of a DNN model $\mathcal{F}_\theta$ to predict a clear image $\hat{\mathbf{x}}$ given a noisy image $\mathbf{y}$ and fixed random noise $\mathbf{z}$ as its input.  
Optimizing the model parameters, it has been shown that the implicit regularization obtained by its structure (DIP), as well as the added explicit TV regularization (DIP-TV), enable the prediction of visually pleasing natural images in low SNRs. 
Using $\mathbf{\Phi}\left( \hat{\mathbf{x}},\mathbf{y} \right)=\|\hat{\mathbf{x}} - \mathbf{y}\|_2^2$, where $\hat{\mathbf{x}} = \mathcal{F}_\theta(\mathbf{z})$ is the model output, our used cost function takes the form:
\begin{equation} \label{eq:image}
    \mathcal{L}_{\text{dif}}\left( \hat{\mathbf{x}},\mathbf{y} \right) = \|\hat{\mathbf{x}} - \mathbf{y}\|_2^2 + \mathcal{L}^T_{\text{sm}}(\hat{\mathbf{x}};\lambda,\rho)
\end{equation}    
We optimize using both our unrolled cost and TV.
Replacing TV regularization with our unrolled smoothness constraint, our method achieved improved results, both quantitatively and qualitatively.

\subsubsection{Implementation}
\input{tables/image_results_summarized}

We remain consistent with the scheme described in \cite{ulyanov2018deep,liu2019image}. 
A U-Net \cite{ronneberger2015u} shaped CNN model with convolution layers added to its skip connections is selected. Its parameters are optimized using an Adam optimizer, given a noisy image and random noise as its input, which remain fixed during training. In each scenario, a noisy image is generated using Additive White Gaussian Noise (AWGN) with variance $\sigma^2$.
The used test data is a combination of 14 grayscale and color images commonly used in image denoising baselines (see Appendix B, available online). Our cost function consists of a data term as described in eq. (\ref{eq:image}), as well as a smoothness regularizer. We compare images reconstructed using standard TV and our smoothness constraint in each experiment. Further implementation details are given in Appendix B (available online).
\input{figures/fig_occ_baselines}
\input{tables/l1vsUnrolled_unified}

\subsubsection{Results}
Results averaged over our used test set are given in table \ref{ta:denoising} (a detailed collection of results for each scenario is provided in Appendix B (available online).
Furthermore, results of two example scenarios are given in table \ref{ta:denoising_spec} and figure \ref{fig:denoising} (further qualitative results are given in Appendix B, available online).
Our method produces improved results on the test set, particularly at low SNRs. 
Moreover, our method demonstrated improved restoration capabilities of edges in piece-wise smooth regions (see examples given in fig. \ref{fig:denoising}). 
In order to quantify this, we measure the Structural Similarity coefficient \cite{wang2004image} between original and reconstructed images, specifically on selected image patches of interest (pSSIM). 
The used patches and corresponding pSSIM coefficients are highlighted in colored bounding boxes (fig. \ref{fig:denoising}) and summarized in table \ref{ta:denoising_spec}.
In conclusion, our method demonstrates improved PSNR particularly in low SNRs, and improved image boundaries preservation,increasing pSSIM by up to 10\%.

\subsection{Unsupervised Optical Flow} \label{sec:of}
\input{tables/sota_all}

\input{figures/fig_l1vsunroll}

We finally test our proposed method in the real-world unsupervised optical flow problem. 
We introduce our smoothness constraint to two well known recent unsupervised optical flow baselines, ARFlow \cite{liu2020learning} and SMURF \cite{stone2021smurf}, featuring two different flow backbones considered highly impactful. Here we demonstrate that replacing TV regularization with our unrolled smoothness constraint during training produces improved flow predictions of both models, particularly at the occluded regions.


Optical flow is defined as a mapping $\mathbf{F}:\mathbb{R}^2 \mapsto \mathbb{R}^2$, which aligns pixels given a pair of images $I_1,I_2$. In the unsupervised optical flow setting, our data term takes the form $\mathbf{\Phi}\left( \mathbf{F},\mathcal{I} \right) = \varphi\left( I_1,\hat{I}_2^{\mathbf{F}} \right)$, where $ֿ\varphi$ is a differentiable photometric loss, measuring image consistency between a reference image and a target image warped using $\mathbf{F}$.  In order to remain consistent with previous optical flow baselines (including ARFlow and SMURF), masked TV regularization is used, replacing $\nabla\mathbf{F}$ with $\mathbf{C=W\odot\nabla F}$ in equation (\ref{eq:loss}) and line 4 in algorithm  \nolinebreak \ref{alg:unrolling}, where $\mathbf{W}$ is a deterministic importance matrix, aiming to decrease the penalty on object boundaries, and $\odot$ stands for point-wise multiplication\footnote{Point-wise multiplication is done through broadcasting the channels of $\mathbf{W}$ to corresponding channels of $\nabla \mathbf{F}=\left[\frac{\partial  u}{\partial x},\frac{\partial  u}{\partial y},\frac{\partial  v}{\partial x},\frac{\partial  v}{\partial y} \right]$.}. $\mathbf{W}$ is defined as
\begin{equation} \label{eq:mask}
\begin{aligned}
    \mathbf{W}&=\exp\left\{-\alpha|\nabla I_1|\right\} \\
    &=\exp\left\{-\alpha\left[\left|\frac{\partial  I_1}{\partial x}\right|,\left|\frac{\partial I_1}{\partial y} \right|\right]\right\}
\end{aligned}
\end{equation}
where $I_1$ is a reference image and $\alpha$ is a hyperparameter controlling the masking operation. In summary, our cost function used for unsupervised optical flow training is:
\begin{equation}
\begin{aligned}
    \mathcal{L}_{\text{dif}}\left(\mathbf{F},\mathcal{I}\right) = \varphi\left( I_1,\hat{I}_2^{\mathbf{F}} \right) +  \mathcal{L}_\text{sm}^T(\mathbf{F};\lambda,\rho) \label{eq:data_flow}
\end{aligned}
\end{equation}

\subsubsection{Model Architectures}
We test our method on two well known recent flow backbones, the light weight Pyramid, Warping, Cost volume (PWC-Net) \cite{Sun2018PWC-Net} and Recurrent All-pairs Field Transforms (RAFT) \cite{teed2020raft}, featured in the ARFlow \cite{liu2020learning} and SMURF \cite{stone2021smurf} baselines, respectively. 

The PWC-Net backbone, featured in the ARFlow baseline, roughly consists of a multi-scale feature extraction pyramid and a Warp, Correlation, Flow (WCF) unit which is shared across all pyramid levels. At each pyramid level, a coarse flow estimate from a higher level is upsampled and fed to the WCF unit, together with the corresponding level features to produce a finer flow estimate. 
The RAFT backbone, featured in the SMURF baseline, consists of feature and context extractors, a 4D cost volume capturing multi-scale all-to-all feature correlations, and a recurrent GRU-based update unit. The update unit performs iterative refinement of the predicted flow until convergence. We refer the reader to \cite{liu2020learning,stone2021smurf} for further details. 


\subsubsection{Datasets and Training} \label{datasets}
We train and evaluate our method on three well-known optical flow benchmarks: the synthetic Flying Chairs \cite{dosovitskiy2015flownet} and MPI Sintel \cite{Butler:ECCV:2012}, and real-world autonomous driving KITTI 2015 \cite{Menze2015CVPR}. 
We rigorously follow the unsupervised training schemes of both ARFlow and SMURF baselines, consisting of pre-training and finetuning steps. Both baselines propose further finetuning in a multi-frame setting, which we do not carry. 

Pre-training our ARFlow model for the MPI Sintel benchmark is done using the raw Sintel movie, consisting on $12,466$ extracted image pairs, split according to scene shots. For the KITTI 2015 benchmark, we pre-train our ARFlow model on the KITTI raw dataset \cite{Geiger2013IJRR}, consisting of real road scenes captured by a car-mounted stereo camera rig. Our SMURF model is pre-trained on the 'training' portion of the Flying Chairs dataset, for both benchmarks. No data labels are used in any step of our training. Occluded pixels are masked out during the evaluation of the data term in eq. \nolinebreak (\ref{eq:data_flow}) following the method suggested in \cite{meister2017unflow} (unsupervised). Thus, occluded regions are learned using the smoothness constraint alone.

Finetuning both our ARFlow and SMURF models for the MPI Sintel benchmark is done using the official training set, consisting of $1,041$ image pairs, each rendered with $2$ levels of difficulty: the clean pass featuring shading only, and the final pass featuring also motion blur, defocus blur and atmospheric effects.
For the KITTI 2015 benchmark, we use the KITTI 2015 multi-view extension for finetuning our models. We exclude frames related to validation from our training set, i.e. use only frames below 9 or above 12 in each scene. Further details regarding the train-test data splitting and training scheme used in our experiments are given in Appendix C (available online). 

As carried in both ARFlow and SMURF baselines, self-supervision from pseudo-random augmentations is acquired during finetuning. Image and corresponding flow augmentations denoted
\begin{subequations} 
\begin{align}
    \mathcal{T}^{\text{img}}:I_k &\mapsto \tilde{I}_k \\ 
    \mathcal{T}^{\text{flow}}: \mathbf{F} &\mapsto \tilde{\mathbf{F}}
\end{align}
\end{subequations} 
perform a combination of appearance, spatial and occlusion pseudo-random deformations. An initial flow estimate $\mathbf{F}_1$ is computed given a pair of images $I_1,I_2$. The image pair then undergoes an image augmentation $\mathcal{T}^{\text{img}}$ resulting in a deformed pair of images $\tilde{I}_1,\tilde{I}_2$. A second flow estimate $\mathbf{F}_2$ is generated next over the deformed images $\tilde{I}_1,\tilde{I}_2$ and a deformed flow $\mathbf{\tilde{F}}_1$ is obtained through applying the corresponding flow augmentation $\mathcal{T}^{\text{flow}}$ over $\mathbf{F}_1$. 
Finally, a self-supervision term measuring the error $\mathbf{F}_2-\mathbf{\tilde{F}}_1$ is constructed and added to the cost function in equation (\ref{eq:data_flow}) during finetuning. Further implementation details are given in Appendix C (available online).

\subsubsection{Flow in Occluded Regions}
\input{figures/sup_flow_iters}
\input{figures/fig_ablations_T_2}

Occluded pixels, i.e. pixels with no correspondence, are highly correlated with motion boundaries. Although crucial to many CV applications, performance on the occluded regions minorly affects the errors averaged over full images. 
These regions are of great interest in this work as they are majorly affected by the gradients of the smoothness constraint during training. We conduct several experiments measuring performance particularly in the occluded regions. 
We train our ARFlow and SMURF models on both the MPI-Sintel and KITTI 2015 data, using our smoothness constraint as well as standard TV (similar to ARFlow, SMURF baselines) and the $L^1$ relaxation methods described at the beginning of section \nolinebreak \ref{sec:exp}.
Measured validation AEPE rates\footnote{Reported validation results are obtained following the train-test data split scheme described in section \ref{datasets}. 
We do not carry multi-frame finetuning in our experiments, as opposed to ARFlow-MV and SMURF official benchmark results} of our unrolled cost, as well as the standard TV, Huber and Charbonnier methods are given in table \nolinebreak \ref{ta:l1vsunrolled}. A qualitative example is given in figure \ref{fig:occ_baselines}. Our method outperforms all tested baselines. As expected, modifying the smoothness regularizer has little effect on the performance in the non-occluded regions (\emph{Noc.}). However, measuring the performance on the occluded regions, our method reduces TV EPE rates by up to 15.82\% enabling the detection of sharper motion boundaries with no added complexity as is highly visible in figure \ref{fig:occ_baselines}.
\input{tables/admm_iterations}
\input{tables/complexity_all}

\subsubsection{Comparison with State of the Art}
We compare our trained ARFlow model with recently published unsupervised methods on the MPI Sintel and KITTI 2015 optical flow benchmarks in table \ref{ta:sota_all}. 
Qualitative examples are given in figure \ref{fig:l1vsunroll}.
The used measures are the Average End-Point-Error (AEPE) and F1 score, measuring outliers percentage (error $>$ 3px). 
All reported MPI Sintel results are of models which used the entire official training data for both training and validation. KITTI 2015 results either share our training scheme of using the multi-view extension excluding frames related to validation, train using the KITTI raw dataset only \cite{meister2017unflow} or use the entire multi-view extension for training, including the image pairs used for validation \cite{jonschkowski2020matters}. 
Comparing to the ARFlow \cite{liu2020learning} and UFlow \cite{jonschkowski2020matters} baselines is particularly interesting, as both are recently published baselines featuring a PWC-Net backbone over dual images only (like ours).
Our method managed to improve the reported results of the ARFlow \cite{liu2020learning} baseline on both MPI Sintel and the KITTI 2015 unsupervised optical flow benchmarks, particularly at the motion boundaries, as is highly visible in all qualitative examples (figures \ref{fig:teaser}, \ref{fig:occ_baselines} and \ref{fig:l1vsunroll}). Moreover, our method reports the best results using a PWC-Net based backbone on dual images at the time of submission.

\subsubsection{Number of Cost Function Update Steps} \label{sec:exp_steps}
We study the effects of varying $T$ values, both on the accumulated gradients during training and on the performance of the trained model during inference.
Cost function update terms $\ell^{(t)}$ and norms of sparsified flow gradients $\mathbf{Q}^{(t)}$, recorded during training on both MPI Sintel and KITTI \nolinebreak 2015, are given in figure \ref{fig:sup_iters}. 
It is evident that both terms converge at $T=2$, and that update terms generated at $T>2$ have little effect on the gradients generated during training. This realization justifies our truncation in equation \nolinebreak (\ref{eq:loss_ours}).
Furthermore, we perform several experiments on the MPI Sintel data (see Appendix C.1.4, available online, for train-test data subsplit). 
We use an ARFlow model which has been pretrained on the MPI Sintel raw movie setting $T=2$, but set various $T$ values during finetuning\footnote{Results reported in this table were obtained through appearance and spatial augmentations during finetuning.}. The results of chosen $T$ values are compared in table \ref{ta:admm_iters} and a qualitative example is shown in figure \ref{fig:ab_T}. As expected, we find that increasing $T$ enables capturing finer motion details. However, it can be seen here as well that the gain for setting $T>2$ is relatively small.
We conclude that setting $T=2$ is optimal as it both achieves good performance and enables preserving computational complexity during training, as is elaborated next (in \ref{sec:exp_comp}).

\subsubsection{Computational Complexity} \label{sec:exp_comp}
Preserving computational complexity during training is achieved through the following. Firstly, we compute our smoothness term at the flow original resolution, which is 1/4 or 1/8 image resolution for ARFlow or SMURF, respectively (this is commonly done in flow baselines), as opposed to upsampling prior to TV computation. Secondly, we limit $T=2$ update terms in our smoothness constraint as explained earlier. 
Our method does not increase model size, as no learned parameters are added. As our method deals with model training alone, it does not affects inference at all. We compare both memory consumption and training time of our method with standard TV in table \ref{ta:comp}. 
Measurements are carried using a single RTX2080 GPU. We use a batch size of 4 image pairs and an image size of $384\times832$ for the ARFlow baseline, and a batch of 1 image pair and image size of $368\times496$ for SMURF. Under these settings, we conclude that our method produces the results described above, while consuming slightly less memory and preserving training time.

%% file: tables/toy_results.tex
\begin{table}
\begin{center}
\begin{tabular}{l c c} 
    \toprule
    \textbf{Method} & \textbf{Final Grad Norm} & \textbf{Pred. Error} \\
    \midrule
    Standard TV     & $3.81 \pm 0.06 \times10^{-5}$   & $2.24 \pm 0.79\times10^{-2}$\\
    Charbonnier    & $1.20 \pm 0.71 \times10^{-8}$     & $1.67 \pm 0.64 \times10^{-2}$\\
    Huber           & $3.77 \pm 2.81 \times10^{-9}$     & $1.62 \pm 0.65 \times10^{-2}$\\    
    \textbf{Ours}   & $\mathbf{7.82 \pm 0.79 \times10^{-10}}$     & $\mathbf{1.40 \pm 0.03 \times10^{-2}}$\\
    \bottomrule
\end{tabular}
\end{center}
\caption{\textbf{PC signal prediction - measured results.} Presented are the measured (best) gradient norms and prediction errors of models trained using our unrolled cost, standard TV regularization and TV relaxations.}
\label{ta:toy}
\end{table}

%% file: figures/fig_img_denoising_all.tex
\begin{figure*}
\begin{center}
\includegraphics[width=1\linewidth]{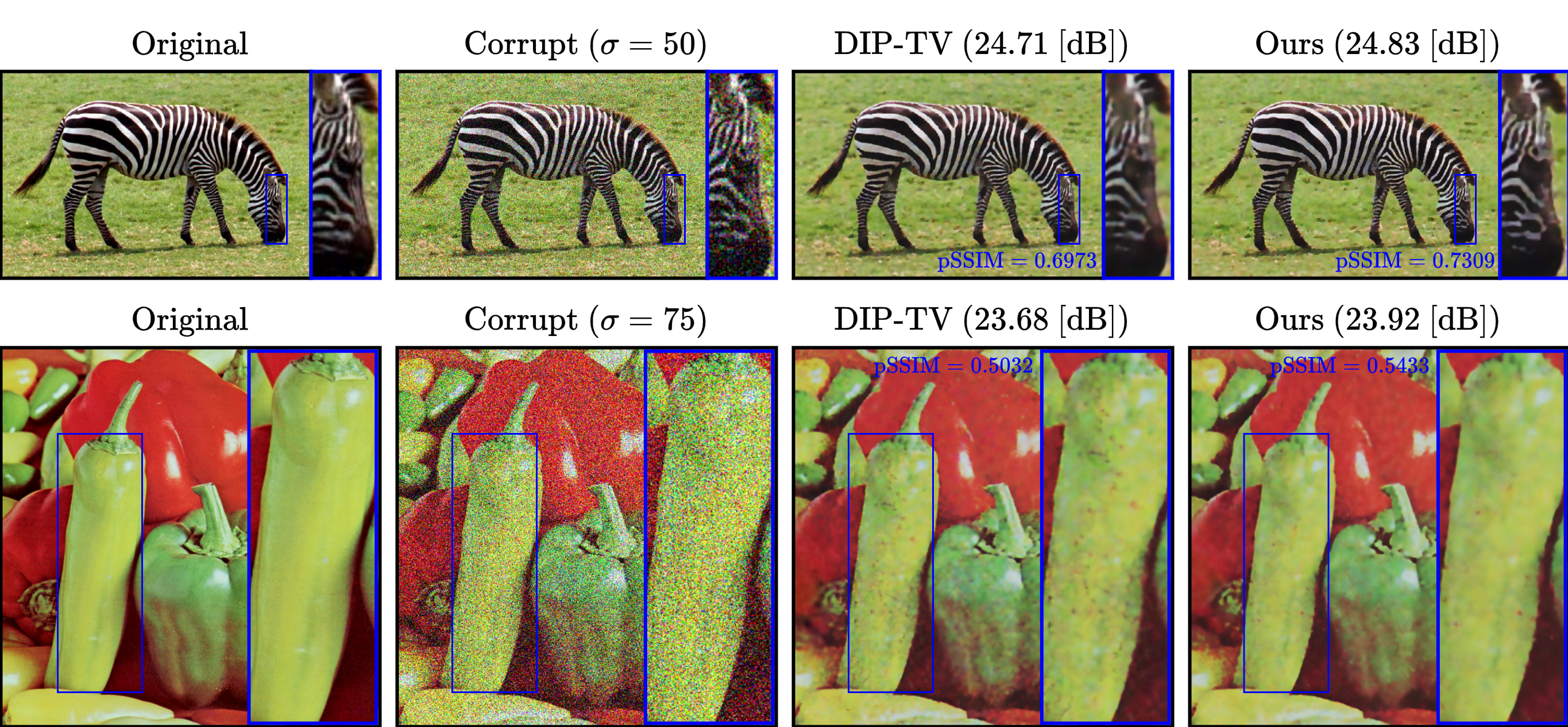}
   \caption{\textbf{Image denoising qualitative examples.}
   Displayed are images corresponding to two example scenarios - \emph{Zebra} (top) and \emph{Peppers} (bottom). From left to right: original, corrupt and reconstructed images obtained by TV regularization (DIP-TV) and our method. The corrupt images are labeled with the added noise levels, and each reconstruction is labeled with its corresponding PSNR [dB] with respect to the original image. Patches and corresponding pSSIM scores are highlighted in colored bounding boxes.
    }
   \label{fig:denoising}
\end{center}
\end{figure*}

%% file: tables/image_results_spec.tex
\begin{table}
\begin{center}
\begin{tabular*}{\linewidth}{@{\extracolsep{\fill}} l | c | l | c | c}
    \toprule
    \textbf{Image} & 
    $\boldsymbol{\sigma}$ & 
    \textbf{Method} & 
    \textbf{PSNR [dB]} &
    \textbf{pSSIM} \emph{(blue)}\\
    \midrule
    \multirow{2}{*}{\emph{Zebra}}    & \multirow{2}{*}{50}    & DIP-TV        & 24.71 & 0.6973 \\
    &  & \textbf{Ours} & \textbf{24.83} & \textbf{0.7309} \\
    \midrule
    \multirow{2}{*}{\emph{Peppers}}  & \multirow{2}{*}{75}    & DIP-TV        & 23.68 & 0.5032 \\
    &  & \textbf{Ours} & \textbf{23.92} & \textbf{0.5433} \\
    \bottomrule
\end{tabular*}
\end{center}
\caption{\textbf{Image denoising qualitative examples results.} 
Both PSNR [dB] and pSSIM scores for selected patches of interest are summarized for the two scenarios given in fig. \ref{fig:denoising}. The added noise level is described by $\sigma$. 
}
\label{ta:denoising_spec}
\end{table}

%% file: tables/image_results_summarized.tex
\begin{table}
\begin{center}
\begin{tabular*}{\linewidth}{@{\extracolsep{\fill}} l | c c c c }
    \toprule
    \multirow{2}{*}{\textbf{Method}} & 
    \multicolumn{4}{c}{\textbf{Ave. PSNR [dB] vs. Noise Levels}} \\
     & $\sigma=25$  & $\sigma=50$ & $\sigma=75$ & $\sigma=100$ \\
    \midrule
    DIP-TV & $27.91$ & $24.72$ & $22.16$ & $20.13$ \\
    \midrule
    \textbf{Ours}    & $\mathbf{27.94}$ & $\mathbf{24.73}$ & $\mathbf{22.22}$ & $\mathbf{20.22}$ \\
    \bottomrule
\end{tabular*}
\end{center}
\caption{\textbf{Summarized image denoising results.} 
PSNRs measured between original and reconstructed images are averaged over the used test set, for each noise level.
The measured average SNR levels (corresponding to $\sigma=25,50,75,100$) are $13.36$, $7.34$, $3.82$ and $1.32$ [dB].
}
\label{ta:denoising}
\end{table}

%% file: figures/fig_occ_baselines.tex
\begin{figure*}
\begin{center}
\includegraphics[width=1\linewidth]{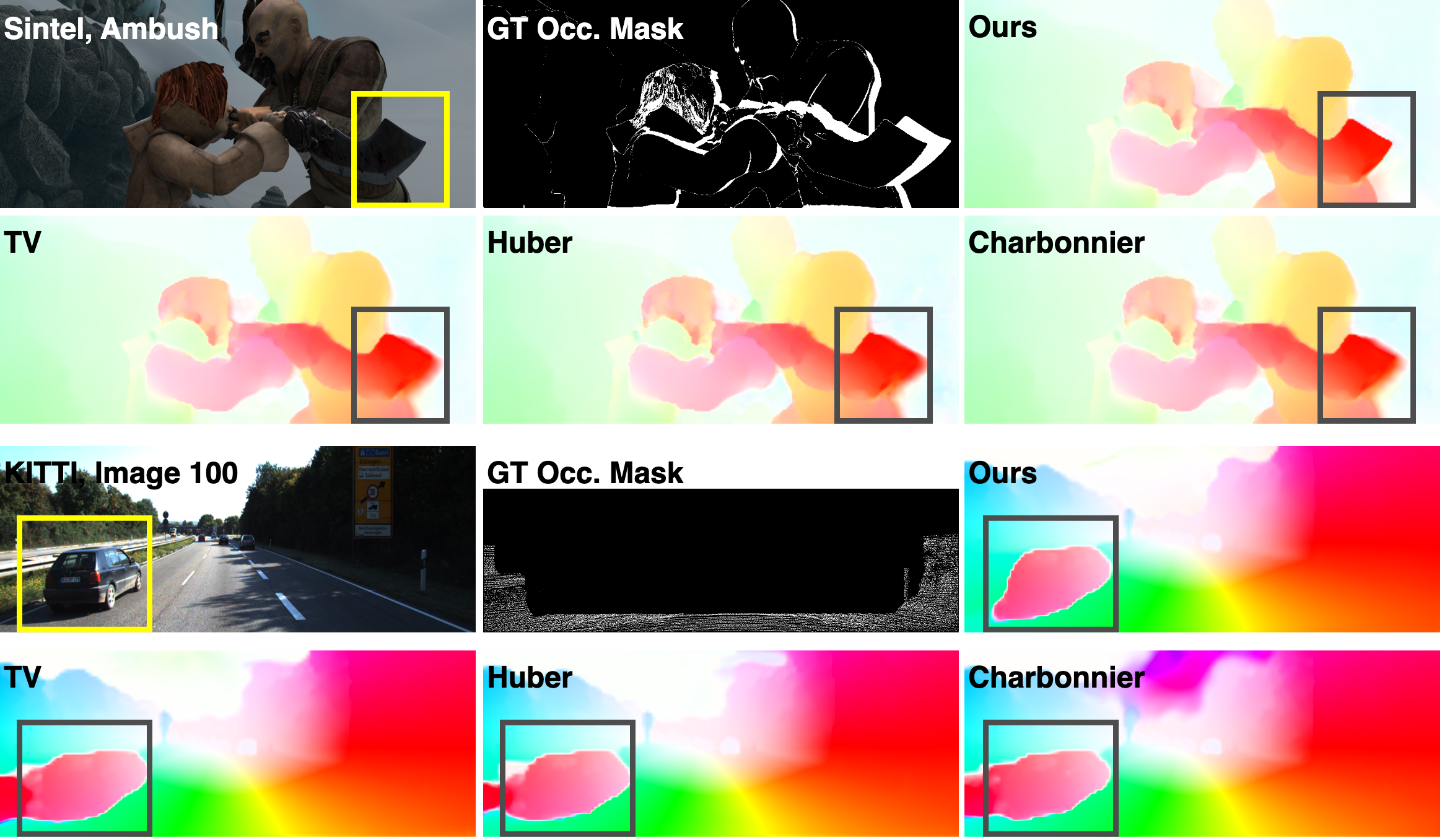}
   \caption{\textbf{Unrolled cost vs. TV relaxations - qualitative examples.}
   Given are Sintel (top) and KITTI (bottom) flows predicted by ARFlow and SMURF baselines, respectively, using our method vs. standard TV, Huber and Charbonnier baselines. The reference images and GT occlusion masks (white stands for occluded) are also given. Training using our smoothness regularizer produces flows which are significantly more accurate at the occluded regions which are correlated with motion boundaries.}
   \label{fig:occ_baselines}
\end{center}
\end{figure*}

%% file: tables/l1vsUnrolled_unified.tex
\begin{table*}
\begin{center}
\begin{tabular*}{.98\linewidth}{@{\extracolsep{\fill}} c|l|ccc|ccc|ccc}
    \toprule
    \multirow{2}{*}{\textbf{Baseline}} & \multirow{2}{*}{\textbf{Method}} & \multicolumn{3}{c}{\textbf{\underline{Sintel Clean}}} & \multicolumn{3}{c}{\textbf{\underline{Sintel Final}}} & \multicolumn{3}{c}{\textbf{\underline{KITTI 2015}}} \\
    & & \textbf{\emph{Occ.}} & \emph{Noc.} & \textbf{\emph{All}} & \textbf{\emph{Occ.}} & \emph{Noc.} & \textbf{\emph{All}} & \textbf{\emph{Occ.}} & \emph{Noc.} & \textbf{\emph{All}} \\
    \midrule
    \multirow{4}{*}{\textbf{ARFlow$^\dagger$}} &
    Standard TV                                     & 8.77 & 1.27 & 2.29 & 9.97 & 2.19 & 3.22 & 6.51 & 2.16 & 2.91\\
   &  Charbonnier \cite{charbonnier1997deterministic} & 8.96 & 1.31 & 2.34 & 9.98 & 2.20 & 3.22 & 6.23 & 2.13 & 2.89\\
    & Huber \cite{huber1964robust}                    & 8.80 & 1.26 & 2.27 & 10.10 & 2.22 & 3.27 & 6.60 & 2.11 & 2.93 \\
    & \textbf{Unrolled Cost (Ours)}  & \textbf{7.91} & 1.28 & \textbf{2.17} & \textbf{9.16} & 2.13 & \textbf{3.06} & \textbf{5.48} & 2.14 & \textbf{2.82}\\
    \midrule
    \multirow{4}{*}{\textbf{SMURF$^\dagger$}} &
    Standard TV                                     & 8.88 & 1.10 & 2.17 & 10.46 & 2.08 & 3.25 & 6.43 & 2.31 & 3.17\\
    & Charbonnier \cite{charbonnier1997deterministic} & 8.89 & 1.09 & 2.17 & 10.48 & 2.04 & 3.23 & 6.63 & 2.40 & 3.2\\
    & Huber \cite{huber1964robust}                    & 8.68 & 1.07 & 2.11 & 10.56 & 2.08 & 3.27 & 6.96 & 2.21 & 3.13 \\
    & \textbf{Unrolled Cost (Ours)}  & \textbf{8.24} & 1.10 & \textbf{2.07} & \textbf{9.88} & 2.11 & \textbf{3.16} & \textbf{5.62} & 2.17 & \textbf{2.89}\\
    \bottomrule
\end{tabular*}
\end{center}
\caption{\textbf{Unrolled cost vs. TV relaxations - unsupervised optical flow.} Our method outperforms all tested $L^1$ relaxations for both tested baselines. Furthermore, our method decreases the TV EPE averaged over the occluded regions, which are highly affected by the smoothness constraint, by up to 15.82\%, enabling the detection of sharper motion boundaries without any modification to the model architecture or complexity. $^\dagger$ All displayed results are ones obtained through our conducted experiments, using the official ARFlow \cite{liu2020learning} and SMURF \cite{stone2021smurf} published code.
}
\label{ta:l1vsunrolled}
\end{table*}

%% file: tables/sota_all.tex
\begin{table*}
\begin{center}
\begin{tabular*}{.8\linewidth}{@{\extracolsep{\fill}}l | c c | c c | c c}
    \toprule
    \multirow{2}{*}{\textbf{Method}} & \multicolumn{2}{c}{\textbf{\underline{Sintel Clean}}}& \multicolumn{2}{c}{\textbf{\underline{Sintel Final}}}  &
    \multicolumn{2}{c}{\textbf{\underline{KITTI 2015}}}\\
    &\emph{Train} & \emph{Test} & \emph{Train} & \emph{Test} & \emph{Train} & \emph{Test (F1)} \\
    \midrule
    CoT-AMFlow $^\star$ \cite{wang2020cot}              & - & 3.96 & - & 5.14 & - & 10.34\% \\
    STFlow $^\star$ \cite{tian2020unsupervised}         & (2.91) & 6.12 & (3.59) & 6.63 & 3.56 & 13.83\% \\
    UPFlow $^\star$ \cite{luo2021upflow}                & (2.33) & 4.68 & (2.67) & 5.32 & 2.45 & 9.38\% \\
    ARFlow-MV $^\dagger$ \cite{liu2020learning}                & (2.73) & 4.49 & (3.69) & 5.67 & (3.46) & 11.79\% \\
    SMURF $^\star$ $^\dagger$ \cite{stone2021smurf}                & (1.71) & 3.15 & (2.58) & 4.18 & (2.00) & 6.83\% \\
    \midrule
    DDFlow \cite{liu2019ddflow}                         & (2.92) & 6.18 & (3.98) & 7.40 & 5.72 & 14.29\% \\
    UFlow-train \cite{jonschkowski2020matters}          & (\textbf{2.50}) & 5.21 & (\textbf{3.39}) & 6.50 & (\textbf{2.71}) & 11.13\% \\
    SimFlow \cite{im2020unsupervised}                   & (2.86) & 5.92 & (3.57) & 6.92 & 5.19 & 13.38\% \\
    ARFlow \cite{liu2020learning}                       & (2.79) & 4.78 & (3.73) & 5.89 & 2.85 & 11.80\% \\
    \textbf{UnrolledCost (Ours)}                        & (2.75) & \textbf{4.69} & (3.61) & \textbf{5.80} & 2.87 & \textbf{10.81\%} \\
    \bottomrule
\end{tabular*}
\end{center}
\caption{\textbf{MPI Sintel \& KITTI 2015 official unsupervised optical flow benchmark results.} We report AEPE and F1 rates for recently published unsupervised methods on both the MPI Sintel \cite{Butler:ECCV:2012} and KITTI 2015 \cite{Menze2015CVPR} optical flow benchmarks. Methods featuring a different (heavier) backbone are marked with $\star$. Multi-frame results are marked with $\dagger$. Brackets "()" indicate results of models trained using their validation set. Best methods in each category are in bold. Missing results are marked as "-". 
By adapting our unrolled cost to the ARFlow \cite{liu2020learning} baseline we not only achieve improved results, but also report the best results for a PWC-Net based backbone on both benchmarks at the time of submission. Moreover, we find our method highly effective on the motion boundaries, as shown in the qualitative results (figures \ref{fig:teaser},\ref{fig:l1vsunroll}).
}
\label{ta:sota_all}
\end{table*}

%% file: figures/fig_l1vsunroll.tex
\begin{figure*}
\begin{center}
\includegraphics[width=1\linewidth]{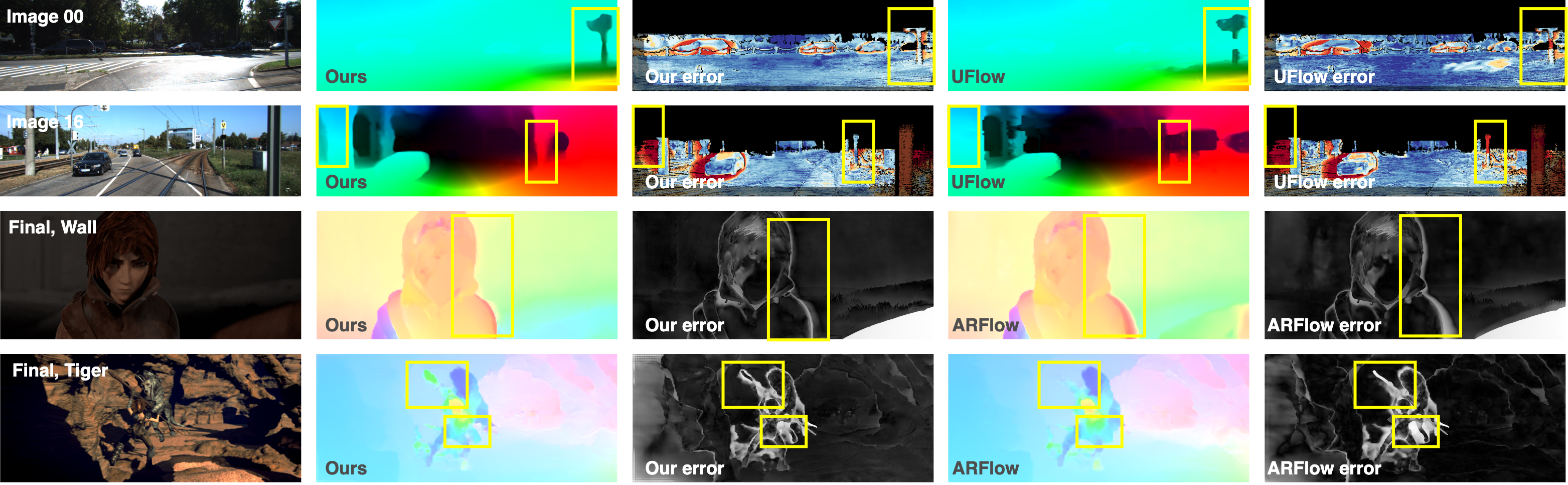}
   \caption{\textbf{Qualitative benchmark results.}
   We compare qualitative flow benchmark results of ours, the ARFlow \cite{liu2020learning} and UFlow \cite{jonschkowski2020matters} baselines. Both ARFlow and UFlow are methods adopting a PWC-Net \cite{Sun2018PWC-Net} based backbone, reporting the best results on the MPI Sintel \cite{Butler:ECCV:2012} and KITTI 2015 \cite{Menze2015CVPR} benchmarks, respectively. 
   We find adapting our unrolled cost to a PWC-Net based backbone outperforms both baselines, particularly at the motion boundaries.}
   \label{fig:l1vsunroll}
\end{center}
\end{figure*}

%% file: figures/sup_flow_iters.tex
\begin{figure}[t]
\begin{center}
\includegraphics[trim= 0 0 10 10, width=1\linewidth]{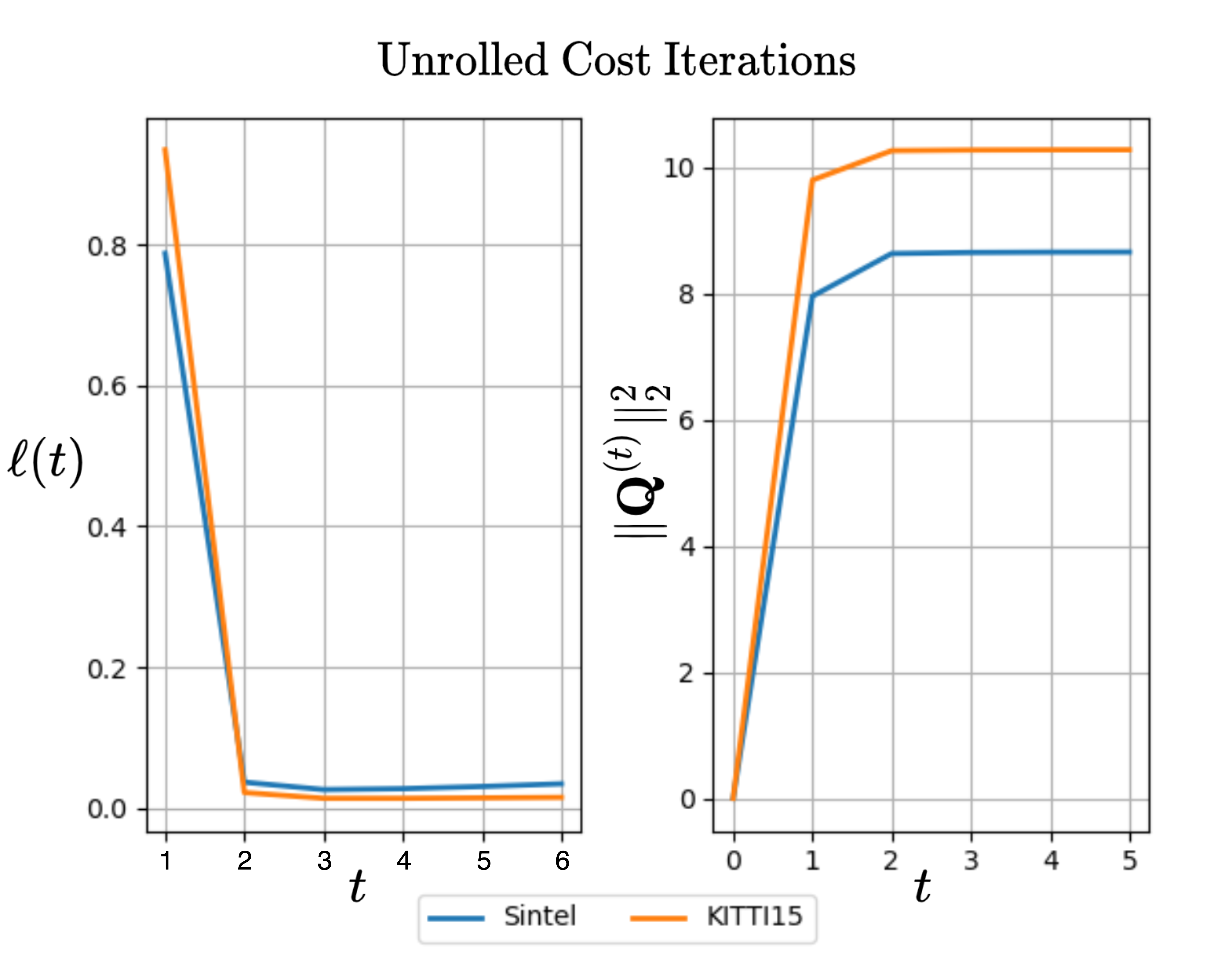}
\end{center}
   \caption{\textbf{Cost function update steps convergence.}
   Displayed are the cost function update steps $\ell^{(t)}$, as well as the $L^2$ norm of the sparsified flow gradients $\mathbf{Q}^{(t)}$, measured during a training iteration on both MPI Sintel and KITTI 2015 datasets.
   }
\label{fig:sup_iters}
\end{figure}

%% file: figures/fig_ablations_T_2.tex
\begin{figure}
\begin{center}
\includegraphics[width=1\linewidth]{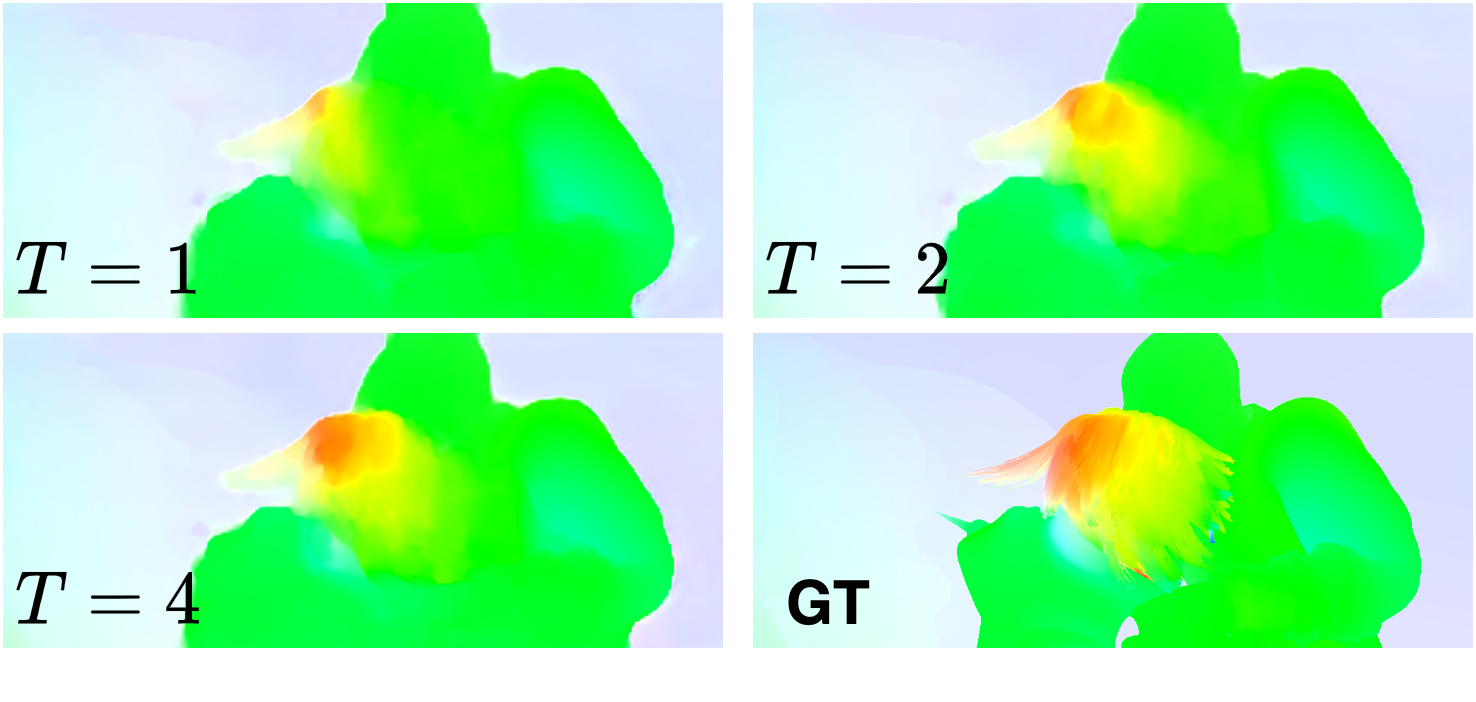}
\end{center}
   \caption{\textbf{Qualitative example - number of update steps.}
   We find increasing the number of update steps improves the ability of our model to capture finer flow details.}
\label{fig:ab_T}
\end{figure}

%% file: tables/admm_iterations.tex
\begin{table}
\begin{center}
\begin{tabular}{l c c} 
    \toprule
    \textbf{$\#$ Update Steps} & \textbf{\underline{Sintel Clean}} & \textbf{\underline{Sintel Final}} \\
    \midrule
    $T=1$               & 2.43   & 3.25 \\
    $T=2$               & 2.35   & 3.19 \\
    $\mathbf{T=4}$      & \textbf{2.35}   & \textbf{3.16} \\
    \bottomrule
\end{tabular}
\end{center}
\caption{\textbf{Number of cost function update steps.} 
    Finetuning using several $T$ values, we find that increasing the number of update steps brings a performance boost.
}
\label{ta:admm_iters}
\end{table}

%% file: tables/complexity_all.tex
\begin{table}
\begin{center}
\begin{tabular}{c|lcc} 
    \toprule
    \textbf{Baseline} & \textbf{Method} & \textbf{Memory} & \textbf{Runtime} \\
    \midrule
    \multirow{2}{*}{\textbf{ARFlow}}    & Standard TV                       & 4.326Gi           & 51ms \\
                                        & \textbf{Unrolled Cost (Ours)}     & \textbf{4.285Gi}  & 51ms \\
    \midrule
    \multirow{2}{*}{\textbf{SMURF}}     & Standard TV                       & 4.86Gi            & 800ms \\
                                        & \textbf{Unrolled Cost (Ours)}     & \textbf{4.80Gi}   & 800ms \\
    \bottomrule
\end{tabular}
\end{center}
\caption{\textbf{Computational complexity.} 
    Our method consumes slightly less memory during training and preserves runtime, hence 
    replacing TV regularization with our unrolled cost enables improved results with no added complexity.
}
\label{ta:comp}
\end{table}

%% file: sections/6_conclusions.tex
\section{Conclusions}
We introduced the concept of Cost Unrolling, improving DNN optimization without changing its architecture or increasing complexity. 
Following our method, we derived a novel smoothness constraint replacing the non-differentiable TV semi norm, commonly used in unsupervised works.
Our derived smoothness constraint is shown to produce more stable gradients during training, thus it enables faster convergence and improved predictions of a given DNN model.
We have demonstrated the effectiveness of our method in a synthetic signal prediction problem as well as image denoising and real-world unsupervised optical flow. Our unrolled cost achieved superior results in all tested scenarios.
We believe that the proposed framework can be applied on top of other model architectures for boosting their results next to non-differentiable optimum solutions.

%% file: bios/gal.tex
\begin{IEEEbiography}[{\includegraphics[width=1in,height=1.25in,clip,keepaspectratio]{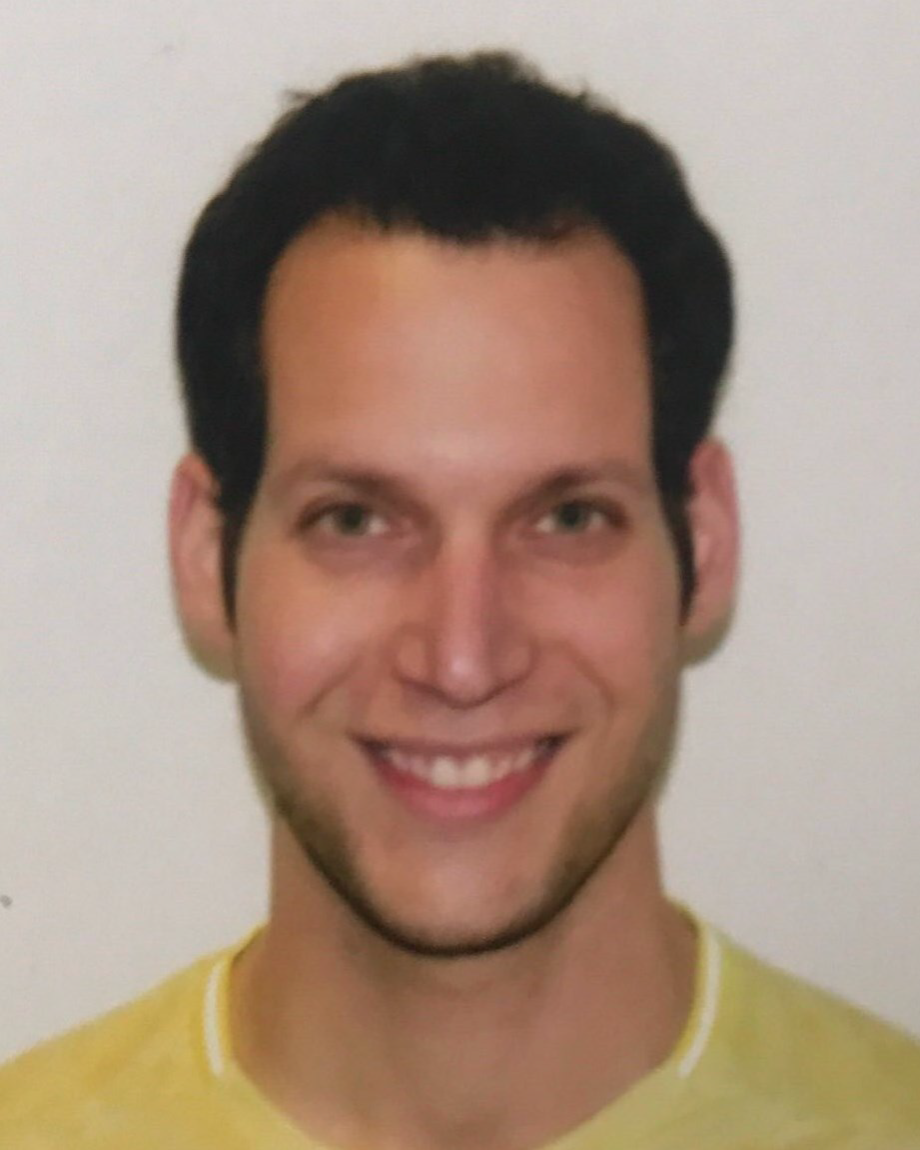}}]{Gal Lifshitz} 
received his B.Sc. and M.Sc. degrees in electrical and computer engineering from Ben-Gurion University, Israel, in 2019, and Tel-Aviv University, Israel, in 2021, respectively.
He is currently pursuing his Ph.D degree in the Geometric Deep Learning lab at Tel-Aviv University, Israel.
His research interests include Computer Vision, Optimization and Statistical Signal Processing.
\end{IEEEbiography}

%% file: bios/dan.tex
\begin{IEEEbiography}[{\includegraphics[width=1in,height=1.25in,clip,keepaspectratio]{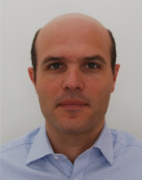}}]{Dr. Dan Raviv} 
is a faculty member in Tel Aviv University, Israel, in the Engineering department, school of Electrical engineering. His work is focused in the intersection between Machine learning, Computer Vision, and non-rigid Geometry. He graduated from the Technion, Israel Institute of Technology, in the Computer Science faculty, where he specialized in metric geometry for non-rigid objects, following several years working in MIT, USA, developing new learnable computer vision models. Dan is the recipient of the prestigious biennial award of SIAM/Imaging Sciences, a graduate of the Technion’s excellence program, and leading the Geometric Deep Learning laboratory in his department.\end{IEEEbiography}

%% file: paper_jrnl_compsoc.bbl
\begin{thebibliography}{10}
\providecommand{\url}[1]{#1}
\csname url@samestyle\endcsname
\providecommand{\newblock}{\relax}
\providecommand{\bibinfo}[2]{#2}
\providecommand{\BIBentrySTDinterwordspacing}{\spaceskip=0pt\relax}
\providecommand{\BIBentryALTinterwordstretchfactor}{4}
\providecommand{\BIBentryALTinterwordspacing}{\spaceskip=\fontdimen2\font plus
\BIBentryALTinterwordstretchfactor\fontdimen3\font minus \fontdimen4\font\relax}
\providecommand{\BIBforeignlanguage}[2]{{%
\expandafter\ifx\csname l@#1\endcsname\relax
\typeout{** WARNING: IEEEtran.bst: No hyphenation pattern has been}%
\typeout{** loaded for the language `#1'. Using the pattern for}%
\typeout{** the default language instead.}%
\else
\language=\csname l@#1\endcsname
\fi
#2}}
\providecommand{\BIBdecl}{\relax}
\BIBdecl

\bibitem{Huber.Wiley.ea1981Robuststatistics}
P.~Huber, J.~Wiley, and W.~InterScience, \emph{{Robust statistics}}.\hskip 1em plus 0.5em minus 0.4em\relax Wiley New York, 1981.

\bibitem{bredies2010total}
K.~Bredies, K.~Kunisch, and T.~Pock, ``Total generalized variation,'' \emph{SIAM Journal on Imaging Sciences}, vol.~3, no.~3, pp. 492--526, 2010.

\bibitem{RUDIN1992259}
\BIBentryALTinterwordspacing
L.~I. Rudin, S.~Osher, and E.~Fatemi, ``Nonlinear total variation based noise removal algorithms,'' \emph{Physica D: Nonlinear Phenomena}, vol.~60, no.~1, pp. 259--268, 1992. [Online]. Available: \url{https://www.sciencedirect.com/science/article/pii/016727899290242F}
\BIBentrySTDinterwordspacing

\bibitem{doi:10.1080/00207160500069904}
\BIBentryALTinterwordspacing
J.~Savage and K.~Chen, ``An improved and accelerated non-linear multigrid method for total-variation denoising,'' \emph{International Journal of Computer Mathematics}, vol.~82, no.~8, pp. 1001--1015, 2005. [Online]. Available: \url{https://doi.org/10.1080/00207160500069904}
\BIBentrySTDinterwordspacing

\bibitem{zach2007duality}
C.~Zach, T.~Pock, and H.~Bischof, ``A duality based approach for realtime tv-l 1 optical flow,'' in \emph{Joint pattern recognition symposium}.\hskip 1em plus 0.5em minus 0.4em\relax Springer, 2007, pp. 214--223.

\bibitem{786990}
J.~Huang and D.~Mumford, ``Statistics of natural images and models,'' in \emph{Proceedings. 1999 IEEE Computer Society Conference on Computer Vision and Pattern Recognition (Cat. No PR00149)}, vol.~1, 1999, pp. 541--547 Vol. 1.

\bibitem{huber1964robust}
P.~J. Huber, ``Robust estimation of a location parameter: Annals mathematics statistics, 35,'' 1964.

\bibitem{charbonnier1997deterministic}
P.~Charbonnier, L.~Blanc-F{\'e}raud, G.~Aubert, and M.~Barlaud, ``Deterministic edge-preserving regularization in computed imaging,'' \emph{IEEE Transactions on image processing}, vol.~6, no.~2, pp. 298--311, 1997.

\bibitem{admm0}
\BIBentryALTinterwordspacing
S.~Boyd, N.~Parikh, E.~Chu, B.~Peleato, and J.~Eckstein, ``Distributed optimization and statistical learning via the alternating direction method of multipliers,'' \emph{Found. Trends Mach. Learn.}, vol.~3, no.~1, p. 1–122, Jan. 2011. [Online]. Available: \url{https://doi.org/10.1561/2200000016}
\BIBentrySTDinterwordspacing

\bibitem{pytorch}
\BIBentryALTinterwordspacing
A.~Paszke, S.~Gross, F.~Massa, A.~Lerer, J.~Bradbury, G.~Chanan, T.~Killeen, Z.~Lin, N.~Gimelshein, L.~Antiga, A.~Desmaison, A.~Kopf, E.~Yang, Z.~DeVito, M.~Raison, A.~Tejani, S.~Chilamkurthy, B.~Steiner, L.~Fang, J.~Bai, and S.~Chintala, ``Pytorch: An imperative style, high-performance deep learning library,'' in \emph{Advances in Neural Information Processing Systems 32}, H.~Wallach, H.~Larochelle, A.~Beygelzimer, F.~d\textquotesingle Alch\'{e}-Buc, E.~Fox, and R.~Garnett, Eds.\hskip 1em plus 0.5em minus 0.4em\relax Curran Associates, Inc., 2019, pp. 8024--8035. [Online]. Available: \url{http://papers.neurips.cc/paper/9015-pytorch-an-imperative-style-high-performance-deep-learning-library.pdf}
\BIBentrySTDinterwordspacing

\bibitem{jia2014caffe}
Y.~Jia, E.~Shelhamer, J.~Donahue, S.~Karayev, J.~Long, R.~Girshick, S.~Guadarrama, and T.~Darrell, ``Caffe: Convolutional architecture for fast feature embedding,'' \emph{arXiv preprint arXiv:1408.5093}, 2014.

\bibitem{tensorflow2015-whitepaper}
\BIBentryALTinterwordspacing
M.~Abadi, A.~Agarwal, P.~Barham, E.~Brevdo, Z.~Chen, C.~Citro, G.~S. Corrado, A.~Davis, J.~Dean, M.~Devin, S.~Ghemawat, I.~Goodfellow, A.~Harp, G.~Irving, M.~Isard, Y.~Jia, R.~Jozefowicz, L.~Kaiser, M.~Kudlur, J.~Levenberg, D.~Man\'{e}, R.~Monga, S.~Moore, D.~Murray, C.~Olah, M.~Schuster, J.~Shlens, B.~Steiner, I.~Sutskever, K.~Talwar, P.~Tucker, V.~Vanhoucke, V.~Vasudevan, F.~Vi\'{e}gas, O.~Vinyals, P.~Warden, M.~Wattenberg, M.~Wicke, Y.~Yu, and X.~Zheng, ``{TensorFlow}: Large-scale machine learning on heterogeneous systems,'' 2015, software available from tensorflow.org. [Online]. Available: \url{https://www.tensorflow.org/}
\BIBentrySTDinterwordspacing

\bibitem{liu2020learning}
L.~Liu, J.~Zhang, R.~He, Y.~Liu, Y.~Wang, Y.~Tai, D.~Luo, C.~Wang, J.~Li, and F.~Huang, ``Learning by analogy: Reliable supervision from transformations for unsupervised optical flow estimation,'' in \emph{IEEE Conference on Computer Vision and Pattern Recognition(CVPR)}, 2020.

\bibitem{wang2016proximal}
S.~Wang, S.~Fidler, and R.~Urtasun, ``Proximal deep structured models,'' in \emph{Proceedings of the 30th International Conference on Neural Information Processing Systems}, ser. NIPS'16.\hskip 1em plus 0.5em minus 0.4em\relax Red Hook, NY, USA: Curran Associates Inc., 2016, p. 865–873.

\bibitem{monga2019algorithm}
V.~Monga, Y.~Li, and Y.~C. Eldar, ``Algorithm unrolling: Interpretable, efficient deep learning for signal and image processing,'' \emph{arXiv preprint arXiv:1912.10557}, 2019.

\bibitem{zhang2020deep}
K.~Zhang, L.~V. Gool, and R.~Timofte, ``Deep unfolding network for image super-resolution,'' in \emph{Proceedings of the IEEE/CVF Conference on Computer Vision and Pattern Recognition}, 2020, pp. 3217--3226.

\bibitem{luo2021upflow}
K.~Luo, C.~Wang, S.~Liu, H.~Fan, J.~Wang, and J.~Sun, ``Upflow: Upsampling pyramid for unsupervised optical flow learning,'' in \emph{Proceedings of the IEEE/CVF Conference on Computer Vision and Pattern Recognition}, 2021, pp. 1045--1054.

\bibitem{wang2020cot}
H.~Wang, R.~Fan, and M.~Liu, ``Cot-amflow: Adaptive modulation network with co-teaching strategy for unsupervised optical flow estimation,'' \emph{arXiv preprint arXiv:2011.02156}, 2020.

\bibitem{stone2021smurf}
A.~Stone, D.~Maurer, A.~Ayvaci, A.~Angelova, and R.~Jonschkowski, ``Smurf: Self-teaching multi-frame unsupervised raft with full-image warping,'' in \emph{Proceedings of the IEEE/CVF Conference on Computer Vision and Pattern Recognition}, 2021, pp. 3887--3896.

\bibitem{Butler:ECCV:2012}
D.~J. Butler, J.~Wulff, G.~B. Stanley, and M.~J. Black, ``A naturalistic open source movie for optical flow evaluation,'' in \emph{European Conf. on Computer Vision (ECCV)}, ser. Part IV, LNCS 7577, {A. Fitzgibbon et al. (Eds.)}, Ed.\hskip 1em plus 0.5em minus 0.4em\relax Springer-Verlag, Oct. 2012, pp. 611--625.

\bibitem{Menze2015CVPR}
M.~Menze and A.~Geiger, ``Object scene flow for autonomous vehicles,'' in \emph{Conference on Computer Vision and Pattern Recognition (CVPR)}, 2015.

\bibitem{Chambolle1997ImageRV}
A.~Chambolle and P.~Lions, ``Image recovery via total variation minimization and related problems,'' \emph{Numerische Mathematik}, vol.~76, pp. 167--188, 1997.

\bibitem{sun2014quantitative}
D.~Sun, S.~Roth, and M.~J. Black, ``A quantitative analysis of current practices in optical flow estimation and the principles behind them,'' \emph{International Journal of Computer Vision}, vol. 106, no.~2, pp. 115--137, 2014.

\bibitem{gou2022multi}
Y.~Gou, P.~Hu, J.~Lv, J.~T. Zhou, and X.~Peng, ``Multi-scale adaptive network for single image denoising,'' \emph{Advances in Neural Information Processing Systems}, vol.~35, pp. 14\,099--14\,112, 2022.

\bibitem{lin2023graph}
Y.~Lin, M.~Yang, J.~Yu, P.~Hu, C.~Zhang, and X.~Peng, ``Graph matching with bi-level noisy correspondence,'' in \emph{Proceedings of the IEEE/CVF International Conference on Computer Vision}, 2023, pp. 23\,362--23\,371.

\bibitem{teed2020raft}
Z.~Teed and J.~Deng, ``Raft: Recurrent all-pairs field transforms for optical flow,'' \emph{arXiv preprint arXiv:2003.12039}, 2020.

\bibitem{meister2017unflow}
S.~Meister, J.~Hur, and S.~Roth, ``Unflow: Unsupervised learning of optical flow with a bidirectional census loss,'' \emph{arXiv preprint arXiv:1711.07837}, 2017.

\bibitem{kim2020unsupervised}
W.~Kim, A.~Kanezaki, and M.~Tanaka, ``Unsupervised learning of image segmentation based on differentiable feature clustering,'' \emph{IEEE Transactions on Image Processing}, vol.~29, pp. 8055--8068, 2020.

\bibitem{jonschkowski2020matters}
R.~Jonschkowski, A.~Stone, J.~T. Barron, A.~Gordon, K.~Konolige, and A.~Angelova, ``What matters in unsupervised optical flow,'' \emph{arXiv preprint arXiv:2006.04902}, 2020.

\bibitem{chen2014insights}
Y.~Chen, R.~Ranftl, and T.~Pock, ``Insights into analysis operator learning: From patch-based sparse models to higher order mrfs,'' \emph{IEEE Transactions on Image Processing}, vol.~23, no.~3, pp. 1060--1072, 2014.

\bibitem{kobler2017variational}
E.~Kobler, T.~Klatzer, K.~Hammernik, and T.~Pock, ``Variational networks: connecting variational methods and deep learning,'' in \emph{Pattern Recognition: 39th German Conference, GCPR 2017, Basel, Switzerland, September 12--15, 2017, Proceedings 39}.\hskip 1em plus 0.5em minus 0.4em\relax Springer, 2017, pp. 281--293.

\bibitem{abdi2016multi}
M.~Abdi and S.~Nahavandi, ``Multi-residual networks: Improving the speed and accuracy of residual networks,'' \emph{arXiv preprint arXiv:1609.05672}, 2016.

\bibitem{5445028}
M.~V. {Afonso}, J.~M. {Bioucas-Dias}, and M.~A.~T. {Figueiredo}, ``Fast image recovery using variable splitting and constrained optimization,'' \emph{IEEE Transactions on Image Processing}, vol.~19, no.~9, pp. 2345--2356, 2010.

\bibitem{chambolle2011first}
A.~Chambolle and T.~Pock, ``A first-order primal-dual algorithm for convex problems with applications to imaging,'' \emph{Journal of mathematical imaging and vision}, vol.~40, no.~1, pp. 120--145, 2011.

\bibitem{sun2016deep}
J.~Sun, H.~Li, Z.~Xu \emph{et~al.}, ``Deep admm-net for compressive sensing mri,'' in \emph{Advances in neural information processing systems}, 2016, pp. 10--18.

\bibitem{dosovitskiy2015flownet}
A.~Dosovitskiy, P.~Fischer, E.~Ilg, P.~Hausser, C.~Hazirbas, V.~Golkov, P.~Van Der~Smagt, D.~Cremers, and T.~Brox, ``Flownet: Learning optical flow with convolutional networks,'' in \emph{Proceedings of the IEEE international conference on computer vision}, 2015, pp. 2758--2766.

\bibitem{peng2022xai}
X.~Peng, Y.~Li, I.~W. Tsang, H.~Zhu, J.~Lv, and J.~T. Zhou, ``Xai beyond classification: Interpretable neural clustering,'' \emph{The Journal of Machine Learning Research}, vol.~23, no.~1, pp. 227--254, 2022.

\bibitem{zhu2018dehazegan}
H.~Zhu, X.~Peng, V.~Chandrasekhar, L.~Li, and J.-H. Lim, ``Dehazegan: When image dehazing meets differential programming.'' in \emph{IJCAI}, 2018, pp. 1234--1240.

\bibitem{ilg2017flownet}
E.~Ilg, N.~Mayer, T.~Saikia, M.~Keuper, A.~Dosovitskiy, and T.~Brox, ``Flownet 2.0: Evolution of optical flow estimation with deep networks,'' in \emph{Proceedings of the IEEE conference on computer vision and pattern recognition}, 2017, pp. 2462--2470.

\bibitem{Sun2018PWC-Net}
D.~Sun, X.~Yang, M.-Y. Liu, and J.~Kautz, ``{PWC-Net}: {CNNs} for optical flow using pyramid, warping, and cost volume,'' 2018.

\bibitem{ulyanov2018deep}
D.~Ulyanov, A.~Vedaldi, and V.~Lempitsky, ``Deep image prior,'' in \emph{Proceedings of the IEEE conference on computer vision and pattern recognition}, 2018, pp. 9446--9454.

\bibitem{liu2019image}
J.~Liu, Y.~Sun, X.~Xu, and U.~S. Kamilov, ``Image restoration using total variation regularized deep image prior,'' in \emph{ICASSP 2019-2019 IEEE International Conference on Acoustics, Speech and Signal Processing (ICASSP)}.\hskip 1em plus 0.5em minus 0.4em\relax Ieee, 2019, pp. 7715--7719.

\bibitem{ronneberger2015u}
O.~Ronneberger, P.~Fischer, and T.~Brox, ``U-net: Convolutional networks for biomedical image segmentation,'' in \emph{International Conference on Medical image computing and computer-assisted intervention}.\hskip 1em plus 0.5em minus 0.4em\relax Springer, 2015, pp. 234--241.

\bibitem{wang2004image}
Z.~Wang, A.~C. Bovik, H.~R. Sheikh, and E.~P. Simoncelli, ``Image quality assessment: from error visibility to structural similarity,'' \emph{IEEE transactions on image processing}, vol.~13, no.~4, pp. 600--612, 2004.

\bibitem{tian2020unsupervised}
L.~Tian, Z.~Tu, D.~Zhang, J.~Liu, B.~Li, and J.~Yuan, ``Unsupervised learning of optical flow with cnn-based non-local filtering,'' \emph{IEEE Transactions on Image Processing}, vol.~29, pp. 8429--8442, 2020.

\bibitem{liu2019ddflow}
P.~Liu, I.~King, M.~R. Lyu, and J.~Xu, ``Ddflow: Learning optical flow with unlabeled data distillation,'' in \emph{Proceedings of the AAAI Conference on Artificial Intelligence}, vol.~33, 2019, pp. 8770--8777.

\bibitem{im2020unsupervised}
W.~Im, T.-K. Kim, and S.-E. Yoon, ``Unsupervised learning of optical flow with deep feature similarity,'' in \emph{European Conference on Computer Vision}.\hskip 1em plus 0.5em minus 0.4em\relax Springer, 2020, pp. 172--188.

\bibitem{Geiger2013IJRR}
A.~Geiger, P.~Lenz, C.~Stiller, and R.~Urtasun, ``Vision meets robotics: The kitti dataset,'' \emph{International Journal of Robotics Research (IJRR)}, 2013.

\end{thebibliography}
